\RequirePackage{snapshot}
\documentclass[twocolumn]{svjour3}

\usepackage{times}
\usepackage{subfig}
\usepackage{multirow}
\usepackage{epsfig}
\usepackage{graphicx}
\usepackage{amsmath}
\usepackage{amssymb}
\usepackage{natbib}
\usepackage[outercaption]{sidecap}  

\usepackage[pagebackref=true,breaklinks=true,letterpaper=true,colorlinks,bookmarks=false,citecolor=blue]{hyperref}

\newcommand{\changed}[1]{#1}

\journalname{International Journal of Computer Vision}

\title{Visualizing Object Detection Features}
\author{Carl Vondrick \and Aditya Khosla \and Hamed Pirsiavash \and Tomasz Malisiewicz \and Antonio Torralba}

\institute{
C. Vondrick, A. Khosla, H. Pirsiavash, A. Torralba \at
Computer Science and Artificial Intelligence Laboratory \\
Massachusetts Institute of Technology\\
Cambridge, MA 02139 USA \\
Email: \texttt{\{vondrick,khosla,hpirsiav,torralba\}@mit.edu}
\and
T.  Malisiewicz \at
vision.ai\\
Cambridge, MA 02139 USA\\
Email: \texttt{tom@vision.ai}
}

\date{Received: date / Accepted: date}

\smartqed

\DeclareMathOperator*{\argmax}{argmax}
\DeclareMathOperator*{\argmin}{argmin}

\begin{document}

\maketitle

\begin{abstract}

We introduce algorithms to visualize feature spaces used by object detectors.
Our method works by inverting a visual feature back to multiple natural images.
We found that these visualizations allow us to analyze object detection systems
in new ways and gain new insight into the detector's failures. For example,
when we visualize the features for high scoring false alarms, we discovered
that, although they are clearly wrong in image space, they do look deceptively
similar to true positives in feature space. This result suggests that many of
these false alarms are caused by our choice of feature space, and supports that
creating a better learning algorithm or building bigger datasets is unlikely to
correct these errors. By visualizing feature spaces, we can gain a more
intuitive understanding of recognition systems.

\end{abstract}

\section{Introduction}

\begin{figure}
\centering

\includegraphics[width=\linewidth]{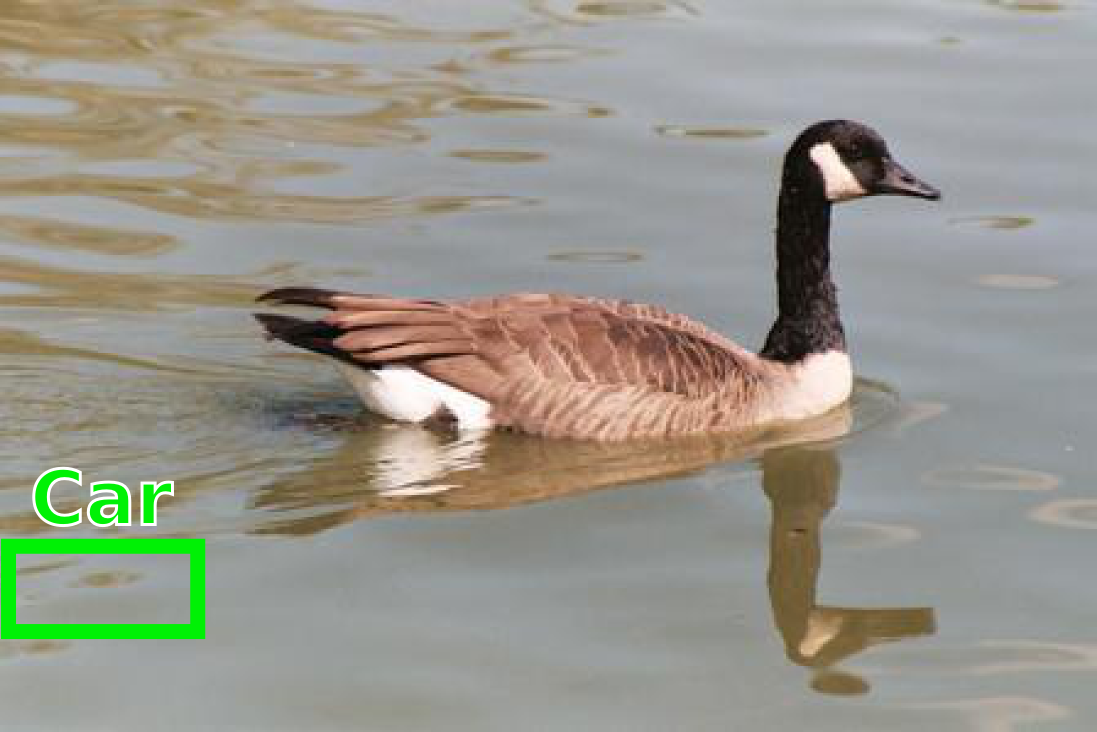}

\vspace{-.5em}
\caption{An image from PASCAL and a high scoring car detection from
DPM \citep{felzenszwalb2010object}. Why did the detector fail?}

\label{fig:teaser}

\vspace{-.5em}
\includegraphics[width=\linewidth]{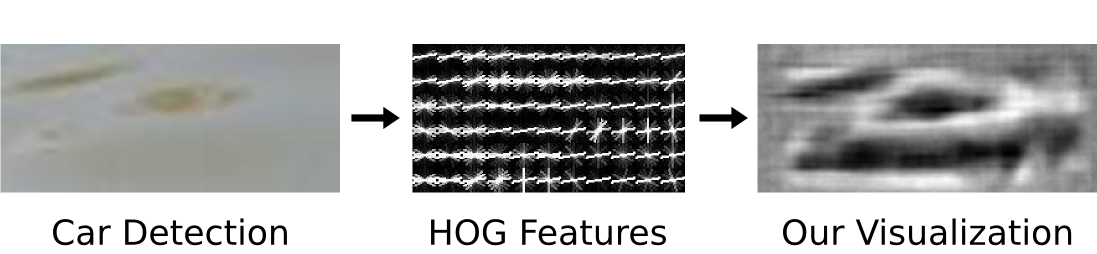}
\vspace{-1em}

\caption{We show the crop for the false car detection from
Figure \ref{fig:teaser}. On the right, we show our visualization of the HOG
features for the same patch. Our visualization reveals that this false alarm
actually looks like a car in HOG space.}

\label{fig:teaser2}

\vspace{-1.5em}

\end{figure}

Figure \ref{fig:teaser} shows a high scoring detection from an object detector
with HOG features and a linear SVM classifier trained on a large database of
images. \changed{\emph{Why} does this detector think that sea water looks like a car?}

\begin{figure*}
\centering

\includegraphics[width=\linewidth]{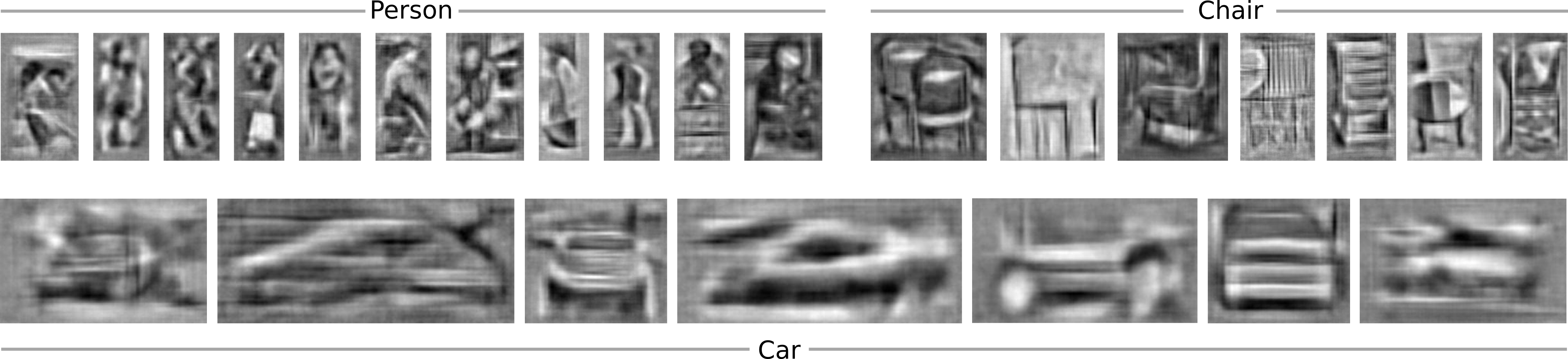}

\caption{We visualize some high scoring detections from the deformable parts
model \citep{felzenszwalb2010object} for person, chair,
and car. Can you guess which are false alarms? Take a minute to study
this figure, then see Figure \ref{fig:topdetsrgb} for the corresponding RGB
patches.}

\label{fig:topdets}

\vspace{-1em}

\end{figure*}

Unfortunately, computer vision researchers are often unable to explain the
failures of object detection systems. Some researchers blame the features,
others the training set, and even more the learning algorithm. Yet, if we wish
to build the next generation of object detectors, it seems crucial to
understand the failures of our current detectors.

In this paper, we introduce a tool to explain some of the failures of object
detection systems. We present algorithms to visualize the
feature spaces of object detectors. Since features are too high dimensional for
humans to directly inspect, our visualization algorithms work by inverting
features back to natural images.  We found that these inversions provide an
intuitive visualization of the feature spaces used by object
detectors.

Figure \ref{fig:teaser2} shows the output from our visualization algorithm on the
features for the false car detection. This visualization reveals that,
while there are clearly no cars in the original image, there is a car hiding
in the HOG descriptor. HOG features see a slightly different visual world than
what we see, and by visualizing this space, we can gain a more intuitive
understanding of our object detectors. 

Figure \ref{fig:topdets} inverts more top detections on PASCAL for a few
categories. Can you guess which are false alarms? Take a minute to study the
figure since the next sentence might ruin the surprise. Although every
visualization looks like a true positive, all of these detections are actually
false alarms.  Consequently, even with a better learning
algorithm or more data, these false alarms will likely persist. In other words,
the features are responsible for these failures. 

\begin{figure}
\includegraphics[width=\linewidth]{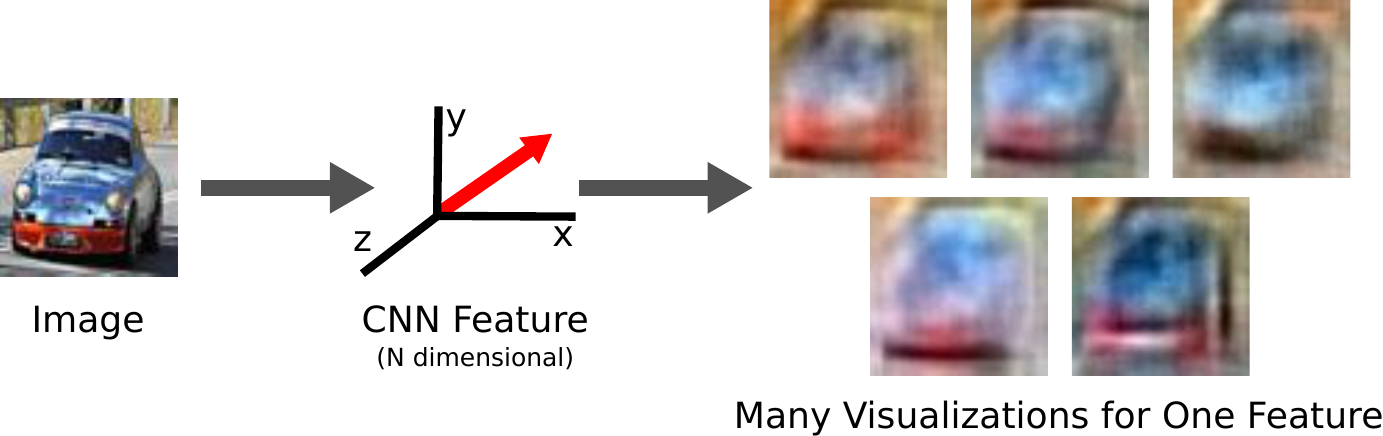}
\caption{Since there are many images that map to similar features, our method recovers multiple images that are diverse in image space, but match closely in feature space.} 
\label{fig:multiple}
\end{figure}

The primary contribution of this paper is a general algorithm for visualizing
features used in object detection. We present a method that inverts visual
features back to images, and show experiments for two standard features in object detection,
HOG and activations from CNNs. Since there are many images that can produce equivalent
feature descriptors, our method moreover recovers multiple images that are perceptually
different in image space, but map to similar feature vectors, illustrated in Figure \ref{fig:multiple}.

\begin{figure}
\includegraphics[width=\linewidth]{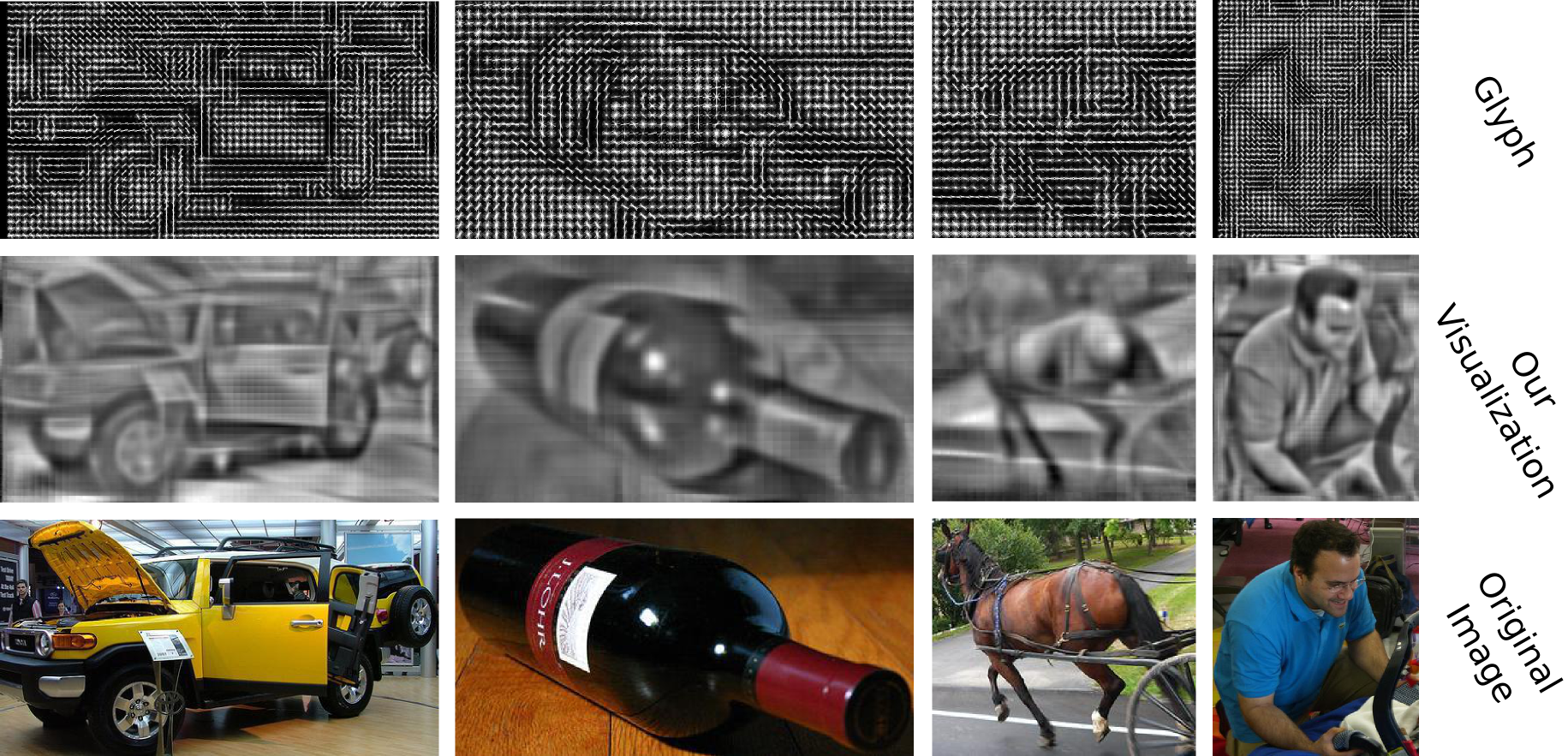}
\caption{In this paper, we present algorithms to visualize features. Our visualizations are more perceptually intuitive for humans to understand.}
\label{fig:example}
\vspace{-1em}
\end{figure}

The remainder of this paper presents and analyzes our visualization algorithm.
We first review a growing body of work in feature visualization for both
handcrafted features and learned representations.  We evaluate our inversions
with both automatic benchmarks and a large human study, and we found our
visualizations are perceptually more accurate at representing the content of a
HOG feature than standard methods; see Figure \ref{fig:example} for a
comparison between our visualization and HOG glyphs. We then use our
visualizations to inspect the behaviors of object detection systems and analyze
their features. Since we hope our visualizations will be useful to other
researchers, our final contribution is a public feature visualization toolbox.\footnote{Available online at \url{http://mit.edu/hoggles}}

\section{Related Work}

Our visualization algorithms are part of an actively growing body of work in
feature inversion.  \cite{oliva2001modeling}, in early work, described a simple
iterative procedure to recover images given gist descriptors.
\citet{weinzaepfel2011reconstructing} were the first to reconstruct an image
given its keypoint SIFT descriptors \citep{lowe1999object}. Their approach
obtains compelling reconstructions using a nearest neighbor based approach on a
massive database. \cite{d2012beyond} then developed an algorithm to reconstruct
images given only LBP features \citep{calonder2010brief,alahi2012freak}.  Their
method analytically solves for the inverse image and does not require a
dataset. \cite{kato2014image} posed feature inversion as a jigsaw puzzle
problem to invert bags of visual words.

Since visual representations that are learned can be difficult to interpret,
there has been recent work to visualize and understand learned features.
\cite{zeiler2013visualizing} present a method to visualize activations from a
convolutional neural network.  In related work, \cite{simonyan2013deep}
visualize class appearance models and their activations for deep networks.
\cite{girshick2013rich} proposed to visualize convolutional neural networks by
finding images that activate a specific feature.  \changed{\cite{deephoggles}
describe a general method for inverting visual features from CNNs by
incorporating natural image priors.}

While these methods are good at reconstructing and visualizing images from
their respective features, our visualization algorithms have some advantages.
Firstly, while most methods are tailored for specific features, the
visualization algorithms we propose are feature independent. Since we cast
feature inversion as a machine learning problem, our algorithms can be used to
visualize any feature. In this paper, we focus on features for object
detection, and we use the same algorithm to invert both HOG and CNN features.
Secondly, our algorithms are fast: our best algorithm can invert features in
under a second on a desktop computer, enabling interactive visualization, which
we believe is important for real-time debugging of vision systems. \changed{Finally, our algorithm
explicitly optimizes for multiple inversions that are diverse in image space, yet match in feature space.}

Our method builds upon work that uses a pair of dictionaries with a coupled
representation for super resolution \citep{yang2010image,wang2012semi} and
image synthesis \citep{huangcoupled}. We extend these methods to show that
similar approaches can visualize features as well. Moreover, we incorporate
novel terms that encourage diversity in the reconstructed image in order
to recover multiple images from a single feature.

Feature visualizations have many applications in computer vision. The computer
vision community has been using these visualization largely to understand object recognition
systems so as to reveal information encoded by features
\citep{zhangspeeding,sadeghi2013fast}, interpret transformations in feature
space \citep{cheninferring}, studying diverse images with similar features
\citep{tatu2011exploring,equiv}, find security failures in machine learning systems
\citep{biggio2012poisoning,weinzaepfel2011reconstructing}, and fix problems in
convolutional neural networks
\citep{zeiler2013visualizing,simonyan2013deep,bruckner2014ml}. With many
applications, feature visualizations are an important tool for the
computer vision researcher.

Visualizations enable analysis that complement a recent line of papers that
provide tools to diagnose object recognition systems, which we briefly review
here.  \cite{parikh2011human,parikh2010role} introduced a new paradigm for
human debugging of object detectors, an idea that we adopt in our experiments.
\cite{hoiem2012diagnosing} performed a large study analyzing the errors that
object detectors make.  \cite{divvala2012important} analyze part-based
detectors to determine which components of object detection systems have the
most impact on performance.  \cite{liu2012has} designed algorithms to highlight
which image regions contribute the most to a classifier's confidence.
\cite{zhuwe} try to determine whether we have reached Bayes risk for HOG.  The
tools in this paper enable an alternative mode to analyze object detectors
through visualizations.  By putting on `HOG glasses' and visualizing the world
according to the features, we are able to gain a better understanding of the
failures and behaviors of our object detection systems.

%The applications of our visualizations extend a recent line of papers that
%examine object detectors and their features. In series of papers, Parikh and
%Zitnick \citep{parikh2011human,parikh2010role} introduced a paradigm for human
%debugging of vision systems.  Hoiem et al.\ \citep{hoiem2012diagnosing}
%performed a large study analyzing the errors that object detectors make.
%Divvala et al.\ \citep{divvala2012important} analyze part-based detectors to
%determine which components of object detection pipelines contribute to the
%performance the most. Tatu et al.\ \citep{tatu2011exploring} explored the set
%of images that generate identical HOG descriptors, and Zhu et al.\
%\citep{zhuwe} try to determine whether we have reached Bayes risk for HOG. The
%tools in this paper complement these previous studies by offering new
%approaches to inspect feature spaces through visualizations. 

%Our most successful algorithm, Paired Dictionary Learning, draws upon recent
%ideas in sparse coding. Where most sparse coding methods focus on learning a
%single dictionary for classification
%\citep{boureau2010learning,lee2007efficient,rendiscriminatively,kavukcuoglu2010learning},
%we extend sparse coding to a regression case and learn a pair of dictionaries
%that allow for efficient transformation between HOG and grayscale images.

\section{Inverting Visual Features} 

We now describe our feature inversion method.  Let $x_0 \in
\mathbb{R}^{P}$ be a natural RGB image and  $\phi = f(x_0) \in \mathbb{R}^Q$ be
its corresponding feature descriptor. \changed{Since features are many-to-one functions,} our goal is to invert the features $\phi$
by recovering a \emph{set} of images $\mathcal{X} = \{x_1, \ldots, x_N\}$ that all map to
the original feature descriptor.

We compute this inversion set $\mathcal{X}$ by solving an optimization problem. We wish
to find several $x_i$ that minimize their reconstruction error in feature space
$\left|\left| f(x_i) - \phi \right|\right|_2^2$ while simultaneously appearing
diverse in image space. We write this optimization as:
\begin{equation}
\begin{aligned}
\mathcal{X} = \argmin_{x, \xi} &\sum_{i=1}^N \left|\left| f(x_i) - \phi \right|\right|_2^2 + \gamma \sum_{j<i} \xi_{ij} \\
\textrm{s.t.} \quad &0 \le S_A(x_i, x_j) \le \xi_{ij} \; \forall_{ij}
\end{aligned}
\label{eqn:objective-multiple}
\end{equation}
The first term of this objective favors images that match in feature space and
the slack variables $\xi_{ij}$ penalize pairs of images that are too similar to
each other in image space \changed{where $S_A(x_i, x_j)$ is the} similarity cost, parametrized by $A$,
between inversions $x_i$ and $x_j$.  A high similarity cost intuitively means
that $x_i$ and $x_j$ look similar and should be penalized. The hyperparameter
$\gamma \in \mathbb{R}$ controls the strength of the similarity cost. By
increasing $\gamma$, the inversions will look more different, at the expense of
matching less in feature space.

\subsection{Similarity Costs}

There are a variety of similarity costs that we could use. \changed{In this work,} we use costs of the form:
\begin{align}
S_A(x_i, x_j) = (x_i^T A x_j)^2 
\label{eqn:similarity}
\end{align}
where $A \in \mathbb{R}^{P \times P}$ is an affinity matrix. Since we are
interested in images that are diverse and not negatives of each other, we
square $x_i^T A x_j$.
The identity affinity matrix, i.e.\ $A = I$, corresponds to comparing inversions
directly in the color space. However, more metrics are also possible, \changed{which we describe now.}

\emph{Edges:} We can design $A$ to favor inversions that differ in edges. Let
$A = C^T C$ where $C \in \mathbb{R}^{2P \times P}$. The first $P$ rows of $C$
correspond to the convolution with the vertical edge filters
$\left[\begin{smallmatrix}-1 & 0 & 1\end{smallmatrix}\right]$ and similarly the
second $P$ rows are for the horizontal edge filters
$\left[\begin{smallmatrix}-1 & 0 & 1\end{smallmatrix}\right]^T$. 

\emph{Color:} We can also encourage the inversions to differ only in colors. 
Let $A = C^T C$ where $C \in \mathbb{R}^{3 \times P}$ is a matrix that averages
each color channel such that $Cx \in \mathbb{R}^3$ is the average RGB color.

\emph{Spatial:} We can force the inversions to only differ in certain spatial
regions. Let $A = C^T C$ where $C \in \mathbb{R}^{P \times P}$ is a binary
diagonal matrix. A spatial region of $x$ will be only encouraged to be
diverse if its corresponding element on the diagonal of $C$ is $1$. Note we can
combine spatial similarity costs with both color and edge costs to encourage
color and edge diversity in only certain spatial regions as well. 

\subsection{Optimization}

Unfortunately, optimizing equation \ref{eqn:objective-multiple} efficiently is challenging because
it is not convex. \changed{Instead, we will make two modifications to solve an approximation:}

\begin{figure}
\includegraphics[width=\linewidth]{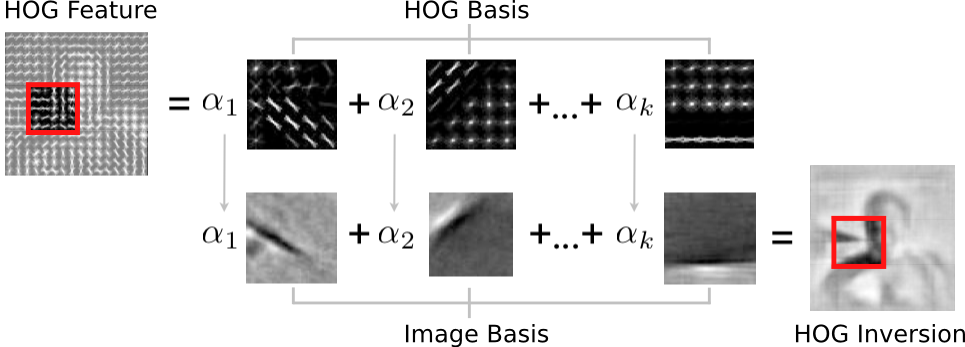}
\caption{Inverting features using a paired dictionary. We first project the feature vector
on to a feature basis. By jointly learning a coupled
basis of features and natural images, we can transfer coefficients estimated from features
to the image basis to recover the natural image.}
\label{fig:pair-tutorial}
\end{figure}

\changed{\emph{Modification 1:} Since the first term of the objective depends on the feature function $f(\cdot)$, which is often not convex nor differentiable, efficient optimization is difficult. Consequently, we approximate an image $x_i$ and its features
$\phi = f(x_i)$ with a paired, over-complete basis to make the objective convex.
Suppose we represent an image $x_i \in \mathbb{R}^P$ and its feature $\phi \in
\mathbb{R}^Q$ in a natural image basis $U \in \mathbb{R}^{P \times K}$ and a
feature space basis $V \in \mathbb{R}^{Q \times K}$ respectively.  We can
estimate $U$ and $V$ such that images and features can be encoded in their
respective bases but with shared coefficients $\alpha \in \mathbb{R}^K$:
\begin{align} x_0 = U\alpha \quad \textrm{and} \quad \phi = V\alpha \end{align}
If $U$ and $V$ have this paired representation, then we can invert features by
estimating an $\alpha$ that reconstructs the feature well.  See Figure
\ref{fig:pair-tutorial} for a graphical representation of the paired
dictionaries.}

\changed{
\emph{Modification 2:} However, the objective is still not convex when there are multiple outputs. We approach solving equation \ref{eqn:objective-multiple} sub-optimally using a greedy approach.
Suppose we already computed the first $i-1$
inversions, $\{x_1, \ldots, x_{i-1}\}$. We then seek the inversion $x_i$ that
is only different from the previous inversions, but still matches $\phi$.
}

Taking these approximations into account, we solve for the inversion $x_i$ with
the optimization:
\begin{equation}
\begin{aligned}
&\alpha_i^* = \argmin_{\alpha_i, \xi} ||V\alpha_i-\phi||_2^2 +  \lambda ||\alpha_i||_1 + \gamma \sum_{j=1}^{i-1} \xi_j \\
& \textrm{s.t.} \quad S_A(U\alpha_i, x_j) \le \xi_j
\end{aligned}
\label{eqn:pair-inverse}
\end{equation}
where there is a sparsity prior on $\alpha_i$ parameterized by $\lambda \in
\mathbb{R}$.\footnote{We found a sparse $\alpha_i$ improves our results. While
our method will work when regularizing with $||\alpha_i||_2$ instead, it tends to
produce more blurred images.} After estimating $\alpha_i^*$, the inversion is
$x_i = U\alpha_i^*$.

\begin{figure}
\includegraphics[width=1\linewidth]{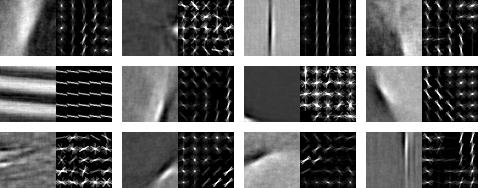}
\caption{Some pairs of dictionaries for $U$ and $V$. The left of every pair is the gray scale dictionary element and the right is the positive components elements in the HOG dictionary. Notice the correlation between dictionaries.}
\label{fig:pair-basis}
\end{figure}

The similarity costs can be seen as adding a weighted Tikhonov 
regularization ($\ell 2$ norm) on $\alpha_i$ because
\begin{align*}
S_A(U \alpha_i, x_j) = \alpha_i^T B \alpha_i \quad
\textrm{where} \quad
B = U^T A^T x_j^T x_j A U
\end{align*}
Since this is combined with lasso, the optimization
behaves as an elastic net \citep{zou2005regularization}.
Note that if we remove the slack variables ($\gamma =
0$), our method reduces to \citep{vondrick2013hog} and only produces one
inversion.

As the similarity costs are in the form of equation \ref{eqn:similarity}, we can
absorb $S_A(x; x_j)$ into the $\ell 2$ norm of equation \ref{eqn:pair-inverse}.
This allows us to efficiently optimize equation \ref{eqn:pair-inverse} using an
off-the-shelf sparse coding solver. We use SPAMS \citep{mairal2009online} in
our experiments. The optimization typically takes a few seconds to produce each
inversion on a desktop computer.

%We greedily apply Eqn.\ref{eqn:pair-inverse} to produce the inversion set $\{x_1, \ldots, x_n\}$. We terminate
%the algorithm once the reconstruction cost $||Va-\phi||_2^2$ exceeds a threshold.

\subsection{Learning}

\begin{figure*}
\centering
\includegraphics[width=\linewidth]{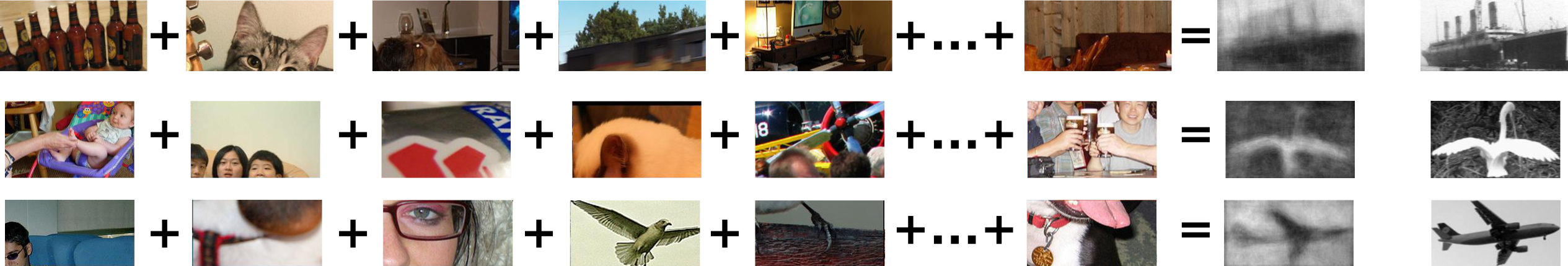}

\caption{We found that averaging the images of top detections from an exemplar
LDA detector provide one method to invert HOG features.}

\label{fig:elda}

\end{figure*}

The bases $U$ and $V$ can be learned such that they have paired coefficients.
We first extract millions of image patches $x_0^{(i)}$ and their corresponding
features $\phi^{(i)}$ from a large database.  Then, we can solve a dictionary
learning problem similar to sparse coding, but with paired dictionaries:
\begin{equation}
\begin{aligned}
\argmin_{U, V, \alpha} \; & \sum_{i} ||x_0^{(i)} - U\alpha_i||_2^2 + ||\phi^{(i)} - V\alpha_i||_2^2 + \lambda ||\alpha_i||_1 \\
\textrm{s.t.} \quad &||U||_2^2 \le \psi_1, \; ||V||_2^2 \le \psi_2 \
\end{aligned}
\label{eqn:pairobj}
\end{equation}
for some hyperparameters $\psi_1 \in \mathbb{R}$ and $\psi_2 \in \mathbb{R}$.
We optimize the above with SPAMS \citep{mairal2009online}. Optimization
typically took a few hours, and only needs to be performed once for a fixed
feature.  See Figure \ref{fig:pair-basis} for a visualization of the learned
dictionary pairs.

\section{Baseline Feature Inversion Methods}

In order to evaluate our method, we also developed several baselines that we use for
comparison. We first describe \changed{three} baselines for single feature inversion, then discuss two
baselines for multiple feature inversion.

\subsection{Exemplar LDA (ELDA)}

Consider the top detections for the exemplar object detector
\citep{hariharan2012discriminative,malisiewicz2011ensemble} for a few
images shown in Figure \ref{fig:elda}. Although all top detections are false
positives, notice that each detection captures some statistics about the query.
Even though the detections are wrong, if we squint, we can see parts of the
original object appear in each detection.

We use this observation to produce our first baseline.
Suppose we wish to invert feature $\phi$. We first train an exemplar LDA
detector \citep{hariharan2012discriminative} for this query,
$w = \Sigma^{-1}(y - \mu)$
where $\Sigma$ and $\mu$ are parameters estimated with a large dataset.
We then score $w$ against every sliding window in this database. The feature
inverse is the average of the top $K$ detections in RGB space:
$f^{-1}(\phi) = \frac{1}{K} \sum_{i=1}^K z_i$ where $z_i$ is an image of a
top detection.

This method, although simple, produces reasonable reconstructions,
even when the database does not contain the category of the feature template.
%We
%note that this method may be subject to dataset bias
%\citep{torralba2011unbiased}. %but could be overcome \citep{khosla2012undoing}.
However, it is computationally expensive since it requires running an object detector across a large database.
\changed{Note that} a similar nearest neighbor method is used in brain research to
visualize what a person might be seeing \citep{nishimoto2011reconstructing}.

\subsection{Ridge Regression}

We describe a fast, parametric
inversion baseline based off ridge regression.
Let $X \in \mathbb{R}^P$ be a random variable representing a gray scale image and $\Phi \in \mathbb{R}^Q$ be a random variable of its corresponding feature. We define
these random variables to be normally distributed on a $P+Q$-variate Gaussian
$P(X, \Phi) \sim \mathcal{N}(\mu, \Sigma)$
with parameters
$\mu = \left[\begin{smallmatrix}
\mu_X & \mu_{\Phi}
\end{smallmatrix}\right]$ and $
\Sigma = \left[\begin{smallmatrix}
\Sigma_{XX} & \Sigma_{X\Phi} \\
\Sigma_{X\Phi}^T & \Sigma_{Y\Phi}
\end{smallmatrix}\right]$.
In order to invert a feature $y$, we calculate the most likely image from
the conditional Gaussian distribution $P(X | \Phi = \phi)$:
\begin{align}
f^{-1}(y) &= \argmax_{x \in \mathbb{R}^D} P(X = x | \Phi = \phi)
\end{align}
It is well known that a Gaussian distribution have a closed form conditional mode: 
\begin{align}
f^{-1}(y) = \Sigma_{X\Phi} \Sigma_{\Phi\Phi}^{-1} (y - \mu_{\Phi}) + \mu_X
\end{align}
Under this inversion algorithm, any feature can be inverted by a single matrix
multiplication, allowing for inversion in under a second. 

We estimate $\mu$ and $\Sigma$ on a large database. In practice, $\Sigma$ is not positive definite; we add
a small uniform prior (i.e., $\hat{\Sigma} = \Sigma + \lambda I$) so $\Sigma$
can be inverted. Since we wish to invert any feature, we assume that $P(X,
\Phi)$ is stationary \citep{hariharan2012discriminative}, allowing us to efficiently learn the covariance across
massive datasets. For features with varying spatial dimensions, we invert a feature by marginalizing
out unused dimensions.

%We found that ridge regression yields blurred inversions.  Intuitively, since
%is invariant to shifts up to its bin size, there are many images that
%map to the same HOG point. Ridge regression is reporting the statistically most
%likely image, which is the average over all shifts. This causes ridge
%regression to only recover the low frequencies of the original image. 

\subsection{Direct Optimization}

We now provide a baseline that attempts to find images that, when we compute features on it, sufficiently match the original descriptor.  In
order to do this efficiently, we only consider images that span a natural image basis.
Let $U \in \mathbb{R}^{D \times K}$ be the natural image basis. We found using the first $K$ eigenvectors of $\Sigma_{XX} \in \mathbb{R}^{D \times D}$ worked well for this basis. Any image $x \in \mathbb{R}^D$ can be encoded by coefficients $\rho \in \mathbb{R}^K$ in this basis: $x = U \rho$. We wish to minimize:
%Since ridge regression only
%recovers the first few principal components of $U$, there is a residual term of high frequencies left to be recovered:
%\begin{align}
%x = \sum_{i=1}^K U \rho_i = \textrm{Low} + \textrm{High} = \phi^{-1}_{B}(y) + \sum_{i=J}^K U \rho_i
%\end{align}
%where $\phi^{-1}_B(\cdot)$ was able to only recover $J$ components. The goal of our third approach is to explicitly recover the high frequency components, i.e. the second term. We wish to minimize:
\begin{equation}
\begin{aligned}
&f^{-1}(y) = U\rho^* \\
&\textrm{where} \quad \rho^* = \argmin_{\rho \in \mathbb{R}^K} \left|\left| f(U\rho) - y \right|\right|_2^2
\end{aligned}
\label{eqn:highfreq-objective}
\end{equation}
Empirically we found success optimizing equation \ref{eqn:highfreq-objective} using
coordinate descent on $\rho$ with random restarts.
We use an over-complete basis corresponding to sparse Gabor-like filters for
$U$. We compute the eigenvectors of $\Sigma_{XX}$ across different scales and
translate smaller eigenvectors to form $U$.

\subsection{Nudged Dictionaries}

In order to compare our ability to recover multiple inversions, we describe two
baselines for multiple feature inversions. 
Our first method modifies paired
dictionaries. Rather than incorporating similarity costs, we add noise to a feature to create a slightly different inversion by ``nudging'' it in random directions:
\begin{equation}
\begin{aligned}
&\alpha_i^* = \argmin_{\alpha_i} ||V\alpha_i-\phi + \gamma \epsilon_i||_2^2 +  \lambda ||\alpha_i||_1\\
\end{aligned}
\end{equation}
where $\epsilon_i \sim \mathcal{N}(0_Q, I_Q)$ is noise from a standard normal distribution \changed{such that
$I_Q$ is the identity matrix}
and $\gamma \in \mathbb{R}$ is a hyperparameter that controls the strength of the diversity.  

\subsection{Subset Dictionaries}
In addition, we compare against a second baseline that
modifies a paired dictionary by removing the basis elements that were
activated on previous iterations. Suppose the
first inversion activated the first $R$ basis elements. We
obtain a second inversion by only giving the paired dictionary the other
$K-R$ basis elements. This forces the sparse coding to use a disjoint basis
set, leading to different inversions.

\section{Evaluation of Single Inversion}

We evaluate our inversion algorithms using both qualitative and quantitative
measures. We use PASCAL VOC 2011 \citep{Everingham10} as our dataset and we
invert patches corresponding to objects. Any algorithm that required training
could only access the training set. During evaluation, only images from the
validation set are examined.  The database for exemplar LDA excluded the
category of the patch we were inverting to reduce the potential effect of
dataset biases. Due to their popularity in object detection, we first focus on
evaluating HOG features.

\begin{figure}
\captionsetup[subfigure]{labelformat=empty}
\centering \subfloat[Original]{ \shortstack{
\includegraphics[width=0.18\linewidth]{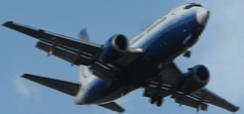} \\
\includegraphics[width=0.18\linewidth]{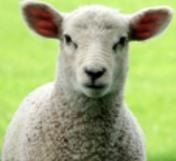} \\
\includegraphics[width=0.18\linewidth]{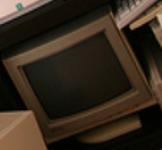}\\
\includegraphics[width=0.18\linewidth]{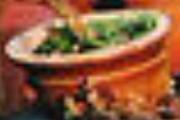} \\
\includegraphics[width=0.18\linewidth]{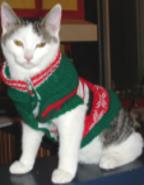} \\
\includegraphics[width=0.18\linewidth]{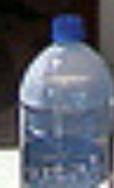} \\
\includegraphics[width=0.18\linewidth]{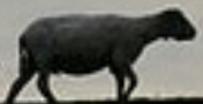}\\
\includegraphics[width=0.18\linewidth]{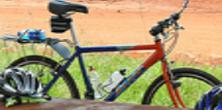} \\
\includegraphics[width=0.18\linewidth]{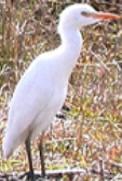} \\
\includegraphics[width=0.18\linewidth]{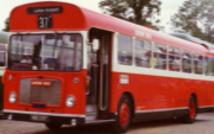} \\
\includegraphics[width=0.18\linewidth]{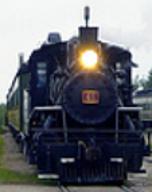}\\
\includegraphics[width=0.18\linewidth]{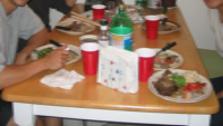}
} }
\subfloat[ELDA]{ \shortstack{
\includegraphics[width=0.18\linewidth]{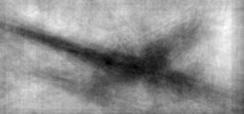} \\
\includegraphics[width=0.18\linewidth]{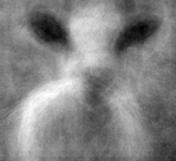} \\
\includegraphics[width=0.18\linewidth]{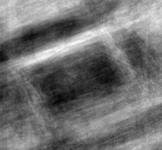} \\
\includegraphics[width=0.18\linewidth]{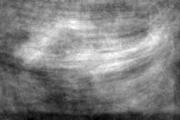} \\
\includegraphics[width=0.18\linewidth]{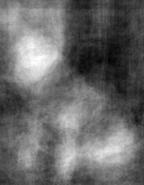} \\
\includegraphics[width=0.18\linewidth]{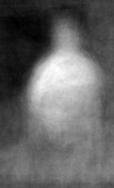} \\
\includegraphics[width=0.18\linewidth]{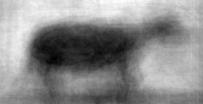} \\
\includegraphics[width=0.18\linewidth]{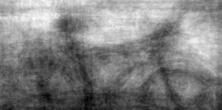} \\
\includegraphics[width=0.18\linewidth]{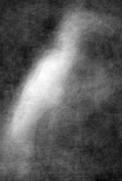} \\
\includegraphics[width=0.18\linewidth]{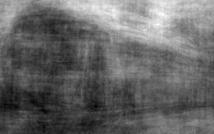} \\
\includegraphics[width=0.18\linewidth]{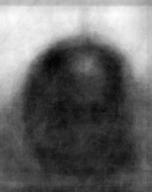}\\
\includegraphics[width=0.18\linewidth]{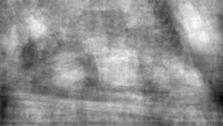}
} }
\subfloat[Ridge]{ \shortstack{
\includegraphics[width=0.18\linewidth]{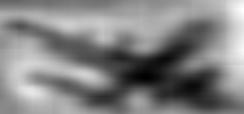} \\
\includegraphics[width=0.18\linewidth]{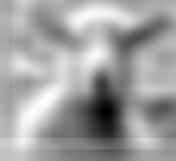}\\
\includegraphics[width=0.18\linewidth]{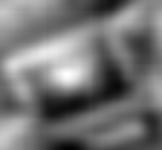} \\
\includegraphics[width=0.18\linewidth]{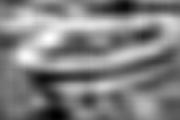} \\
\includegraphics[width=0.18\linewidth]{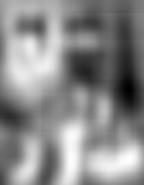} \\
\includegraphics[width=0.18\linewidth]{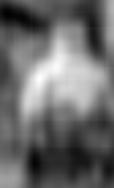} \\
\includegraphics[width=0.18\linewidth]{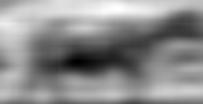} \\
\includegraphics[width=0.18\linewidth]{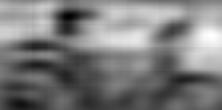} \\
\includegraphics[width=0.18\linewidth]{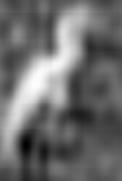} \\
\includegraphics[width=0.18\linewidth]{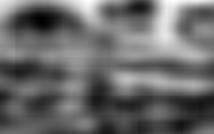}\\
\includegraphics[width=0.18\linewidth]{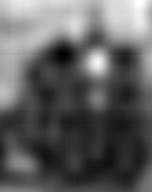}\\
\includegraphics[width=0.18\linewidth]{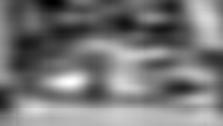}
} }
\subfloat[Direct]{ \shortstack{
\includegraphics[width=0.18\linewidth]{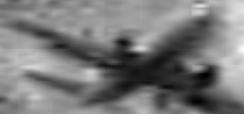} \\
\includegraphics[width=0.18\linewidth]{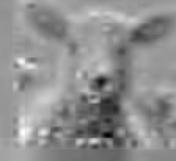} \\
\includegraphics[width=0.18\linewidth]{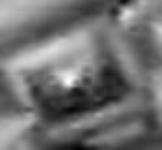} \\
\includegraphics[width=0.18\linewidth]{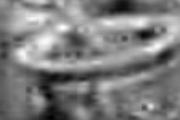}\\
\includegraphics[width=0.18\linewidth]{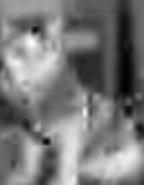} \\
\includegraphics[width=0.18\linewidth]{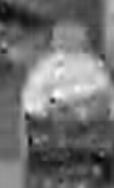} \\
\includegraphics[width=0.18\linewidth]{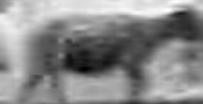}\\
\includegraphics[width=0.18\linewidth]{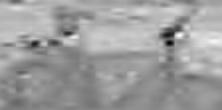}\\
\includegraphics[width=0.18\linewidth]{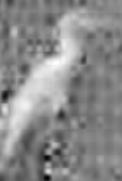} \\
\includegraphics[width=0.18\linewidth]{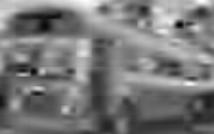}\\
\includegraphics[width=0.18\linewidth]{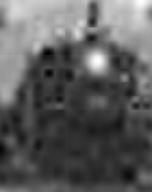}\\
\includegraphics[width=0.18\linewidth]{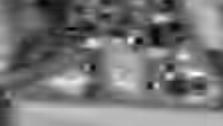}
} }
\subfloat[PairDict]{ \shortstack{
\includegraphics[width=0.18\linewidth]{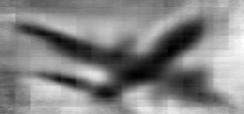} \\
\includegraphics[width=0.18\linewidth]{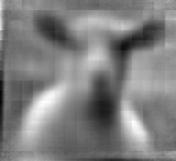} \\
\includegraphics[width=0.18\linewidth]{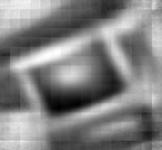} \\
\includegraphics[width=0.18\linewidth]{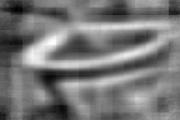}\\
\includegraphics[width=0.18\linewidth]{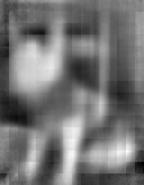} \\
\includegraphics[width=0.18\linewidth]{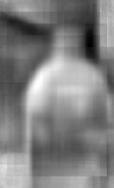} \\
\includegraphics[width=0.18\linewidth]{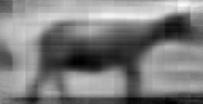}\\
\includegraphics[width=0.18\linewidth]{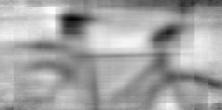}\\
\includegraphics[width=0.18\linewidth]{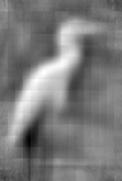} \\
\includegraphics[width=0.18\linewidth]{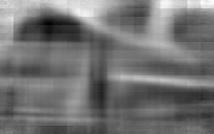}\\
\includegraphics[width=0.18\linewidth]{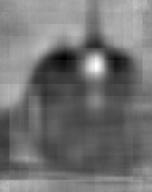}\\
\includegraphics[width=0.18\linewidth]{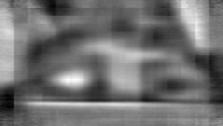}
} }
\caption{We
show results for all four of our inversion algorithms on held out image
patches on similar dimensions common for object detection.}
%In general, exemplar
%LDA produces grainy inversions. Ridge regression is blurry, but fast. 
%Direct optimization is able to recover high frequencies at the expense of extra noise;
%notice the eyes on the sheep and cat, and details on the bus. Paired dictionary learning often
%perceptually performs the best, striking a middle ground between crisp and
%blurry. Please see supplemental material for more results.}
\label{fig:results}

\end{figure}

\subsection{Qualitative Results}

We show our inversions in Figure \ref{fig:results} for a few object categories.
Exemplar LDA and ridge regression tend to produce blurred visualizations.
Direct optimization recovers high frequency details at the expense of extra
noise. Paired dictionary learning tends to produce the best visualization
for HOG descriptors. By learning a dictionary over the visual world and
the correlation between HOG and natural images, paired dictionary learning
recovered high frequencies without introducing significant noise.

\begin{figure*}
\centering
\includegraphics[width=0.45\linewidth]{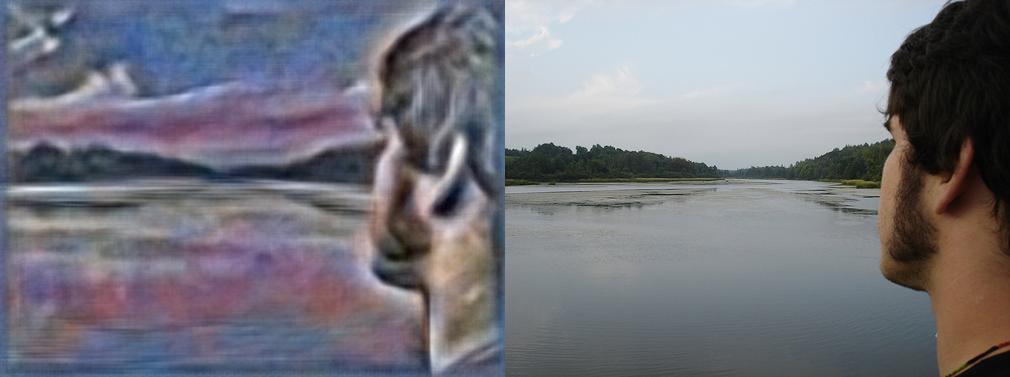}\hspace{2em}\includegraphics[width=0.45\linewidth]{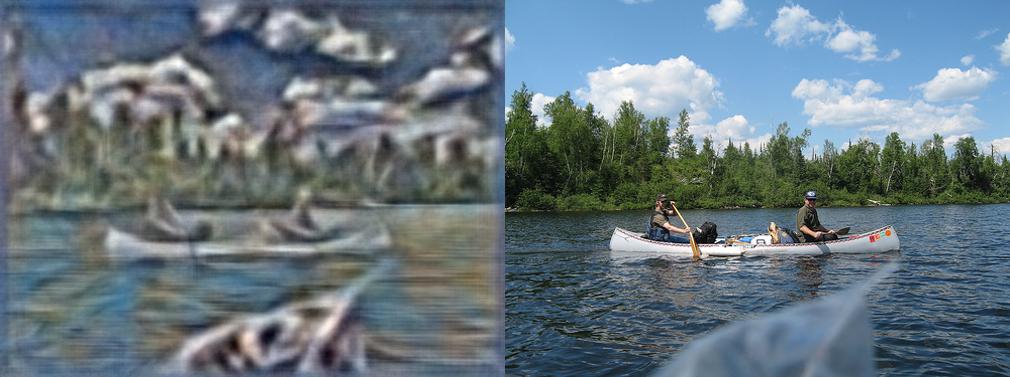}\\
\vspace{0.2em}
\includegraphics[width=0.45\linewidth]{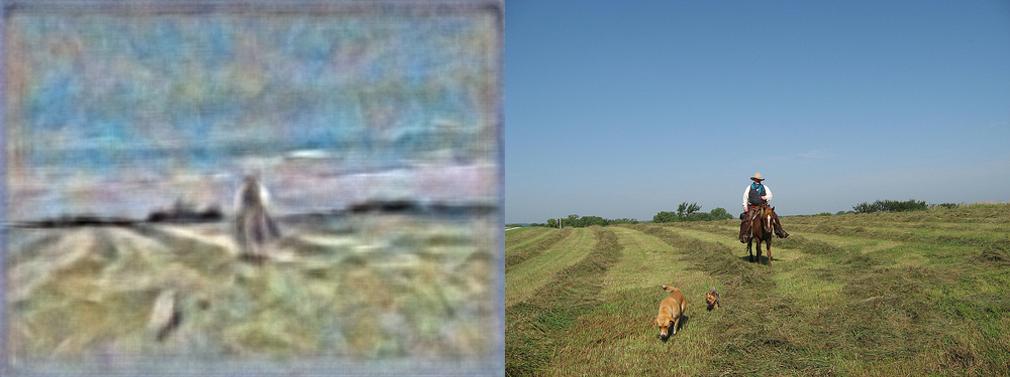}\hspace{2em}\includegraphics[width=0.45\linewidth]{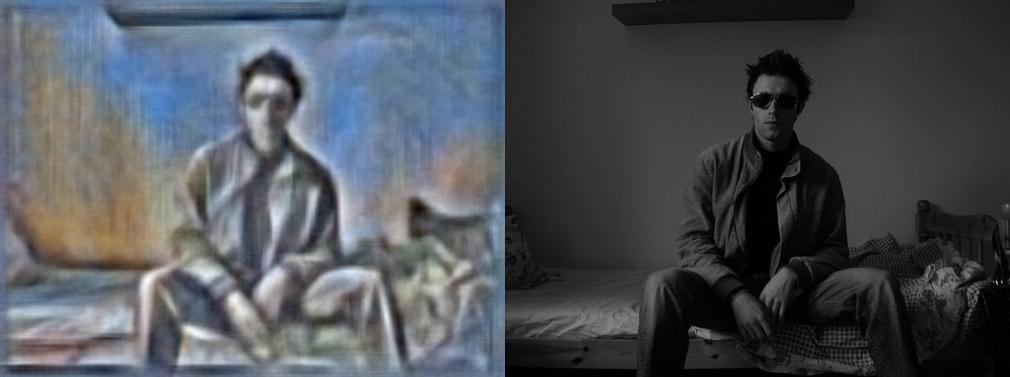}\\
\vspace{0.2em}
\includegraphics[width=0.45\linewidth]{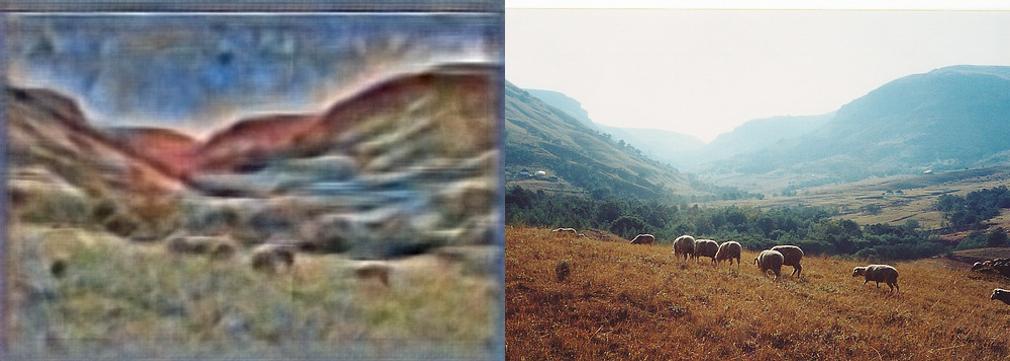}\hspace{2em}\includegraphics[width=0.45\linewidth]{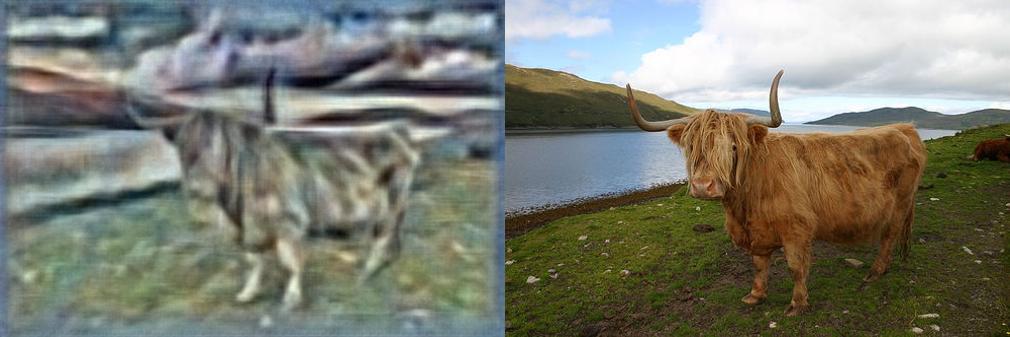}\\
%\includegraphics[width=\linewidth]{figs/color/2011_000807.jpg}
%\vspace{1em}
%\includegraphics[width=0.45\linewidth]{figs/color/2011_000130.jpg}\hspace{2em}\includegraphics[width=0.45\linewidth]{figs/color/2011_001824.jpg}\\
%\includegraphics[width=\linewidth]{figs/color/2011_002453.jpg}
\caption{We show results where our paired dictionary algorithm is trained to
recover RGB images instead of only grayscale images. The right shows the
original image and the left shows the inverse.}
\label{fig:color}
\end{figure*}

Although HOG does not explicitly encode color, we found that the paired
dictionary is able to recover color from HOG descriptors.  Figure
\ref{fig:color} shows the result of training a paired dictionary to estimate
RGB images instead of grayscale images. While the paired dictionary assigns
arbitrary colors to man-made objects and indoor scenes, it frequently colors
natural objects correctly, such as grass or the sky, likely because those
categories are strongly correlated to HOG descriptors. We focus on grayscale
visualizations in this paper because we found those to be more intuitive for
humans to understand.

\begin{figure}
\centering
\includegraphics[width=12em]{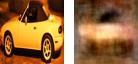}\hspace{2em}
\includegraphics[width=12em]{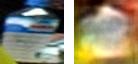}
\includegraphics[width=12em]{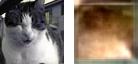}\hspace{2em}
\includegraphics[width=12em]{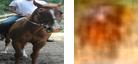}
\includegraphics[width=12em]{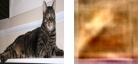}\hspace{2em}
\includegraphics[width=12em]{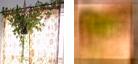}
\includegraphics[width=12em]{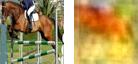}\hspace{2em}
\includegraphics[width=12em]{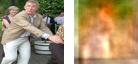}

\caption{\changed{We show visualizations from our method to invert features
from deep convolutional networks. Although the visualizations
are blurry, they capture some key aspects of the original images, such as shapes and colors. Our visualizations are inverting 
the last convolutional layer of \cite{krizhevsky2012imagenet}.}}
\label{fig:qual-icnn}

\end{figure}

\changed{
We also explored whether our visualization algorithm could invert other
features besides HOG, such as deep features. Figure \ref{fig:qual-icnn} shows
how our algorithm can recover some details of the original image given only
activations from the last convolutional layer of \cite{krizhevsky2012imagenet}.
Although the visualizations are blurry, they do capture some important visual
aspects of the original images such as shapes and colors. This suggests that our visualization algorithm
may be general to the type of feature.}

\begin{figure}
\centering
\captionsetup[subfigure]{labelformat=empty}
\subfloat[Original]{
\includegraphics[width=0.32\linewidth]{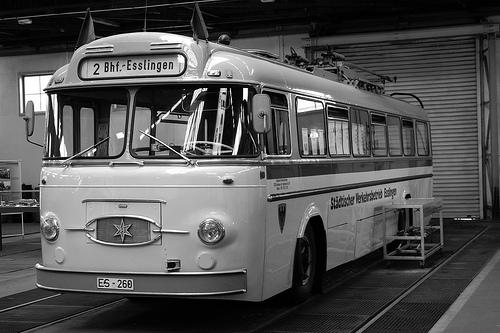}
}
\subfloat[PairDict (seconds)]{
\includegraphics[width=0.32\linewidth]{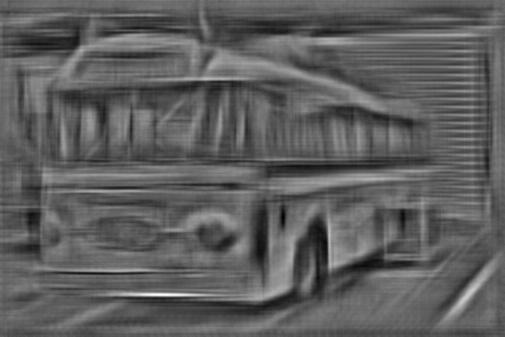}
}
\subfloat[Greedy (days)]{
\includegraphics[width=0.32\linewidth]{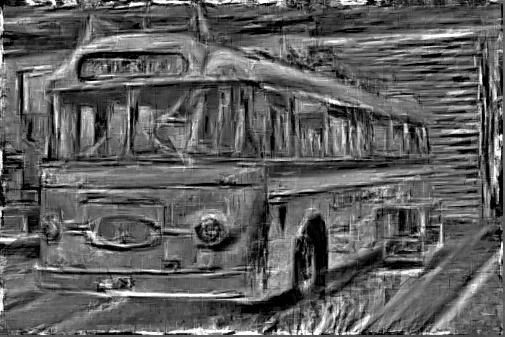}
}
\caption{Although our algorithms are good at inverting HOG, they are not perfect, and struggle to reconstruct high frequency detail. See text for details.}
\label{fig:notperfect}
\end{figure}

\begin{figure}
\captionsetup[subfigure]{labelformat=empty}
\centering
\subfloat[Original $x$]{
\includegraphics[width=0.32\linewidth]{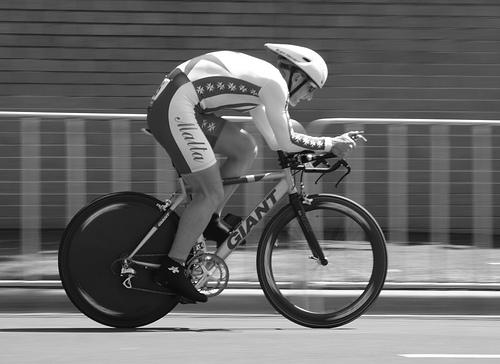}
}
\subfloat[$x' = \phi^{-1}\left(\phi(x)\right)$]{
\includegraphics[width=0.32\linewidth]{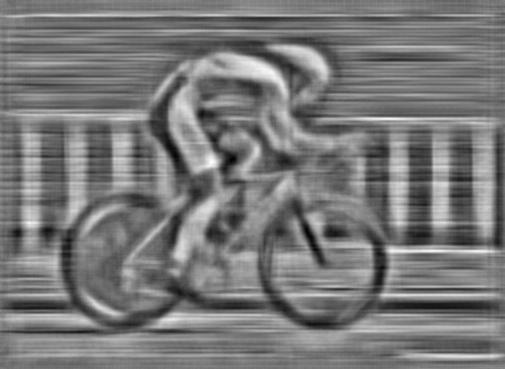}
}
\subfloat[$x'' = \phi^{-1}\left(\phi(x')\right)$]{
\includegraphics[width=0.32\linewidth]{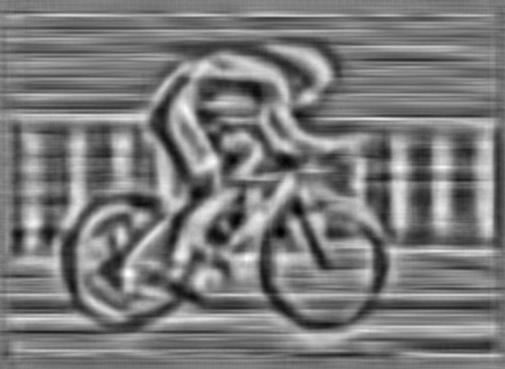}
}
\caption{We recursively compute HOG and invert it with a paired dictionary. While there is some information loss, 
our visualizations still do a good job at accurately representing HOG features. $\phi(\cdot)$ is HOG,  and $\phi^{-1}(\cdot)$ is the inverse.}
\label{fig:recursion}
\end{figure}

\begin{figure}
\centering
\captionsetup[subfigure]{labelformat=empty}
\subfloat[$40 \times 40$]{
\includegraphics[width=0.23\linewidth]{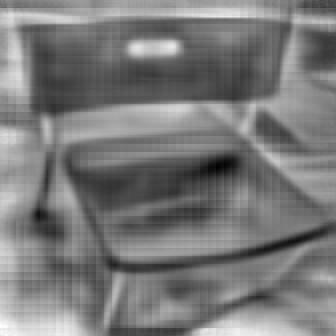}
}
\subfloat[$20 \times 20$]{
\includegraphics[width=0.23\linewidth]{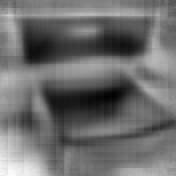}
}
\subfloat[$10 \times 10$]{
\includegraphics[width=0.23\linewidth]{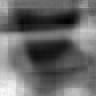}
}
\subfloat[$5 \times 5$]{
\includegraphics[width=0.23\linewidth]{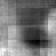}
}
\caption{Our inversion algorithms are sensitive to the HOG template size.
We show how performance degrades as the template becomes smaller.}
\label{fig:pyramid}
\end{figure}

While our visualizations do a good job at representing HOG features, they have
some limitations. Figure \ref{fig:notperfect} compares our best visualization
(paired dictionary) against a greedy algorithm that draws triangles of random
rotation, scale, position, and intensity, and only accepts the triangle if it
improves the reconstruction. If we allow the greedy algorithm to execute for an
extremely long time (a few days), the visualization better shows higher
frequency detail. This reveals that there exists a visualization better than
paired dictionary learning, although it may not be tractable \changed{for large scale experiments}. In a related
experiment, Figure \ref{fig:recursion} recursively computes HOG on the inverse
and inverts it again. This recursion shows that there is some loss between
iterations, although it is minor and appears to discard high frequency details.
Moreover, Figure \ref{fig:pyramid} indicates that our inversions are sensitive
to the dimensionality of the HOG template.  Despite these limitations, our
visualizations are, as we will now show, still perceptually intuitive for
humans to understand.

\subsection{Quantitative Results}

We quantitatively evaluate our algorithms under two benchmarks.  Firstly, we use an automatic
inversion metric that measures how well our inversions reconstruct original
images.  Secondly, we conducted a large visualization challenge with human
subjects on Amazon Mechanical Turk (MTurk), which is designed to determine how
well people can infer high level semantics from our visualizations.

\emph{Pixel Level Reconstruction:}
We consider the inversion performance of our algorithm: given a HOG feature
$y$, how well does our inverse $\phi^{-1}(y)$ reconstruct the original pixels
$x$ for each algorithm? Since HOG is invariant up to a constant shift and
scale, we score each inversion against the original image with normalized cross
correlation. Our results are shown in Table \ref{tab:inversionobjective}.
Overall, exemplar LDA does the best at pixel level reconstruction.

%

%\begin{figure}
%\subfloat[HOG (Us)]{\includegraphics[width=0.33\linewidth]{figs/sift-compare.jpg}}
%\subfloat[Original]{\includegraphics[width=0.33\linewidth]{figs/sift-compare-original.jpg}}
%\subfloat[Keypoint SIFT \citep{weinzaepfel2011reconstructing}]{\includegraphics[width=0.33\linewidth]{figs/sift-compare-theirs.jpg}}
%\caption{We compare our paired dictionary learning approach on HOG with the algorithm of \citep{weinzaepfel2011reconstructing} on SIFT. Our blurred inversion shows that HOG is a more coarse descriptor than keypoint SIFT.}
%\label{fig:siftcompare}
%\end{figure}

\emph{Semantic Reconstruction:}
While the inversion benchmark evaluates how well the inversions reconstruct the
original image, it does not capture the high level content of the inverse: is
the inverse of a sheep still a sheep? To evaluate this, we conducted a study on
MTurk. We sampled 2,000 windows corresponding to objects in
PASCAL VOC 2011. We then showed participants an inversion from one of our
algorithms and asked participants to classify it into one of the 20 categories. Each
window was shown to three different users. Users were required to pass a
training course and qualification exam before participating in order to
guarantee users understood the task. Users could optionally select that they
were not confident in their answer. We also compared our algorithms against the
standard black-and-white HOG glyph popularized by \citep{dalal2005histograms}.

\begin{table}
\centering
\begin{tabular}{l | c c c c c}
Category & ELDA & Ridge & Direct & PairDict \\
\hline
aeroplane &      \textbf{0.634} &  \textbf{0.633} &  0.596 &  0.609 \\
bicycle &        0.452 &  \textbf{0.577} &  0.513 &  0.561 \\
bird &   \textbf{0.680} &  0.650 &  0.618 &  0.638 \\
boat &   \textbf{0.697} &  0.678 &  0.631 &  0.629 \\
bottle &         \textbf{0.697} &  0.683 &  0.660 &  0.671 \\
bus &    0.627 &  \textbf{0.632} &  0.587 &  0.585 \\
car &    0.668 &  \textbf{0.677} &  0.652 &  0.639 \\
cat &    \textbf{0.749} &  0.712 &  0.687 &  0.705 \\
chair &  \textbf{0.660} &  0.621 &  0.604 &  0.617 \\
cow &    \textbf{0.720} &  0.663 &  0.632 &  0.650 \\
table &    \textbf{0.656} &  0.617 &  0.582 &  0.614 \\
dog &    \textbf{0.717} &  0.676 &  0.638 &  0.667 \\
horse &  \textbf{0.686} &  0.633 &  0.586 &  0.635 \\
motorbike &      0.573 &  \textbf{0.617} &  0.549 &  0.592 \\
person &         \textbf{0.696} &  0.667 &  0.646 &  0.646 \\
pottedplant &    \textbf{0.674} &  \textbf{0.679} &  0.629 &  0.649 \\
sheep &  \textbf{0.743} &  0.731 &  0.692 &  0.695 \\
sofa &   \textbf{0.691} &  0.657 &  0.633 &  0.657 \\
train &  \textbf{0.697} &  0.684 &  0.634 &  0.645 \\
tvmonitor &      \textbf{0.711} &  0.640 &  0.638 &  0.629 \\
\hline
Mean & \textbf{0.671} & 0.656  &0.620 & 0.637\\
\end{tabular}

\caption{We evaluate the performance of our inversion algorithm by comparing
the inverse to the ground truth image using the mean normalized cross
correlation. Higher is better; a score of 1 is perfect.}

\label{tab:inversionobjective}

\end{table}

\setlength{\tabcolsep}{2pt}
\begin{table}
\centering
\begin{tabular}{l | c c c c c | c}
Category & ELDA & Ridge & Direct & PairDict & Glyph & Expert \\
\hline
aeroplane & 0.433& 0.391& 0.568& \textbf{0.645}& 0.297 & 0.333\\
bicycle & 0.327& 0.127& 0.362& 0.307& \textbf{0.405} & 0.438 \\
bird & 0.364& 0.263& \textbf{0.378}& 0.372& 0.193 & 0.059\\
boat & 0.292& 0.182& 0.255& \textbf{0.329}& 0.119 & 0.352\\
bottle & 0.269& 0.282& 0.283& \textbf{0.446}& 0.312 & 0.222\\
bus & 0.473& 0.395& \textbf{0.541}& 0\textbf{.549}& 0.122 & 0.118\\
car & 0.397& 0.457& \textbf{0.617}& 0.585& 0.359 & 0.389\\
cat & 0.219& 0.178& \textbf{0.381}& 0.199& 0.139 & 0.286 \\
chair & 0.099& 0.239& 0.223& \textbf{0.386}& 0.119 & 0.167\\
cow & 0.133& 0.103& \textbf{0.230}& 0.197& 0.072 & 0.214\\
table & 0.152& 0.064& 0.162& \textbf{0.237}& 0.071 & 0.125\\
dog & 0.222& 0.316& \textbf{0.351}& 0.343& 0.107 & 0.150\\
horse & 0.260& 0.290& 0.354& \textbf{0.446}& 0.144 & 0.150\\
motorbike & 0.221& 0.232& \textbf{0.396}& 0.224& 0.298 & 0.350\\
person & 0.458& 0.546& 0.502& \textbf{0.676}& 0.301 & 0.375\\
pottedplant & 0.112& 0.109& \textbf{0.203}& 0.091& 0.080 & 0.136\\
sheep & 0.227& 0.194& \textbf{0.368}& 0.253& 0.041 & 0.000\\
sofa & 0.138& 0.100& 0.162& \textbf{0.293}& 0.104 & 0.000\\
train & 0.311& 0.244& 0.316& \textbf{0.404}& 0.173 & 0.133\\
tvmonitor & 0.537& 0.439& 0.449& \textbf{0.682}& 0.354 & 0.666\\
\hline
Mean & 0.282& 0.258& 0.355& \textbf{0.383} & 0.191 & 0.233 
\end{tabular}

\caption{We evaluate visualization performance across twenty PASCAL VOC
categories by asking MTurk participants to classify our inversions.  Numbers are
percent classified correctly; higher is better. Chance is $0.05$.  Glyph refers
to the standard black-and-white HOG diagram popularized by
\citep{dalal2005histograms}.  Paired dictionary learning provides the best
visualizations for humans. Expert refers to MIT PhD students in computer vision
performing the same visualization challenge with HOG glyphs.}

\label{tab:userstudy}
\end{table}

Our results in Table \ref{tab:userstudy} show that paired dictionary learning
and direct optimization provide the best visualization of HOG descriptors for
humans.  Ridge regression and exemplar LDA perform better than the glyph, but
they suffer from blurred inversions.  Human performance on the HOG glyph is
generally poor, and participants were even the slowest at completing that
study. Interestingly, the glyph does the best job at visualizing bicycles,
likely due to their unique circular gradients. Our results overall suggest that
visualizing HOG with the glyph is misleading, and richer visualizations from
our paired dictionary are useful for interpreting HOG features.

%\begin{figure}
%\includegraphics[width=0.5\linewidth,trim=16em 10em 23em 10em]{figs/confusion2.jpg}%\includegraphics[width=0.33\linewidth,trim=16em 10em 23em 10em]{figs/confusion2.jpg}\includegraphics[width=0.33\linewidth,trim=16em 10em 23em 10em]{figs/confusion-expert.jpg}
%\includegraphics[width=0.5\linewidth,trim=16em 10em 23em 10em]{figs/confusion4.jpg}%\includegraphics[width=0.33\linewidth,trim=15em 12em 23em 11em]{figs/confusion3.jpg}\includegraphics[width=0.33\linewidth,trim=15em 12em 23em 11em]{figs/confusion5.jpg}
%\caption{We show the confusion matrices for a paired dictionary and the HOG glyph. The vertical axis is the ground truth category and the horizontal axis is the predicted category. Please see supplemental material for the other confusion matrices.}
%\label{fig:confusionmatrix}
%\end{figure}

%Figure \ref{fig:confusionmatrix} shows the classification confusion matrix for
%our algorithms. Firstly, notice that subjects incorrectly showed a strong prior that the
%inversions were for people, evidenced by a bright vertical bar in the confusion
%matrix. Secondly, notice that subjects tended to make the same confusions
%that object detectors, e.g. confuse bottles with people, or motorbikes 

Our experiments suggest that humans can predict the performance of object detectors
by only looking at HOG visualizations. Human accuracy on
inversions and state-of-the-art object detection AP scores from
\citep{felzenszwalb2010cascade} are correlated with a Spearman's rank
correlation coefficient of 0.77. 

We also asked computer vision PhD students at MIT to classify HOG glyphs in order to
compare MTurk participants with experts in HOG. Our results are summarized
in the last column of Table \ref{tab:userstudy}.  HOG experts performed slightly
better than non-experts on the glyph challenge, but experts on glyphs did not
beat non-experts on other visualizations. This result suggests that our
algorithms produce more intuitive visualizations even for object detection
researchers.

\section{Evaluation of Multiple Inversions}

\changed{Since features are many-to-one functions, our visualization algorithms should be able to recover multiple inversions
for a feature descriptor.
We look at the multiple inversions from deep network features because these features appear to be robust to several invariances.}

To conduct our experiments with multiple inversions, we inverted features from the AlexNet convolutional
neural network \citep{krizhevsky2012imagenet} trained on ImageNet
\citep{deng2009imagenet,russakovsky2014imagenet}.   We use the publicly
available Caffe software package \citep{Jia13caffe} to extract features. We use
features from the last convolutional layer (pool5), which has been shown to
have strong performance on recognition tasks \citep{girshick2013rich}.  We
trained the dictionaries $U$ and $V$ using random windows from the PASCAL VOC
2007 training set \citep{Everingham10}.  We tested on two thousand random
windows corresponding to objects in the held-out PASCAL VOC 2007 validation
set. 

\begin{figure}[t]
\centering
\subfloat[Affinity = Color]{
\includegraphics[width=0.47\linewidth]{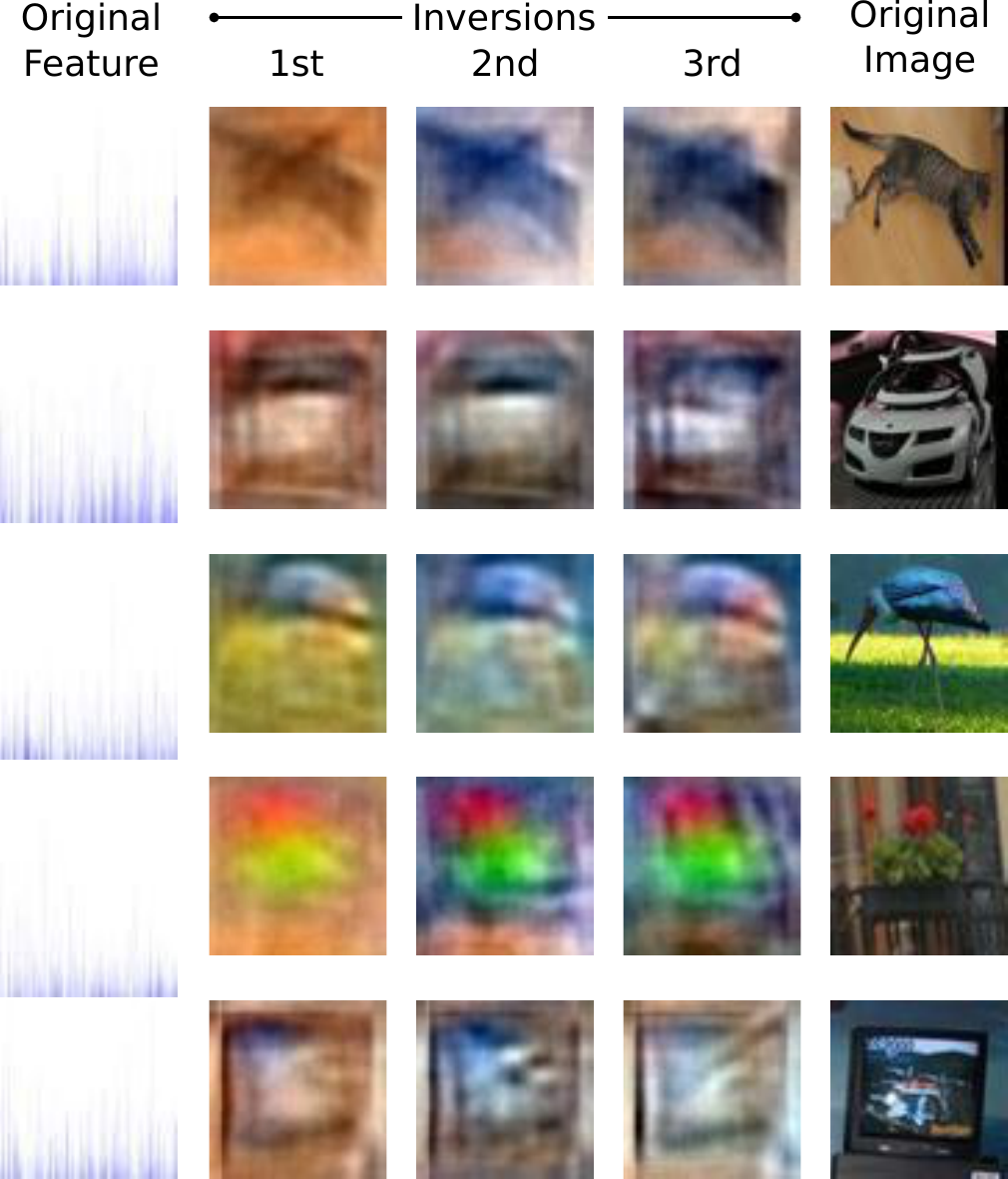}
}
\hspace{0.01\linewidth}
\subfloat[Affinity = Edge]{
\includegraphics[width=0.47\linewidth]{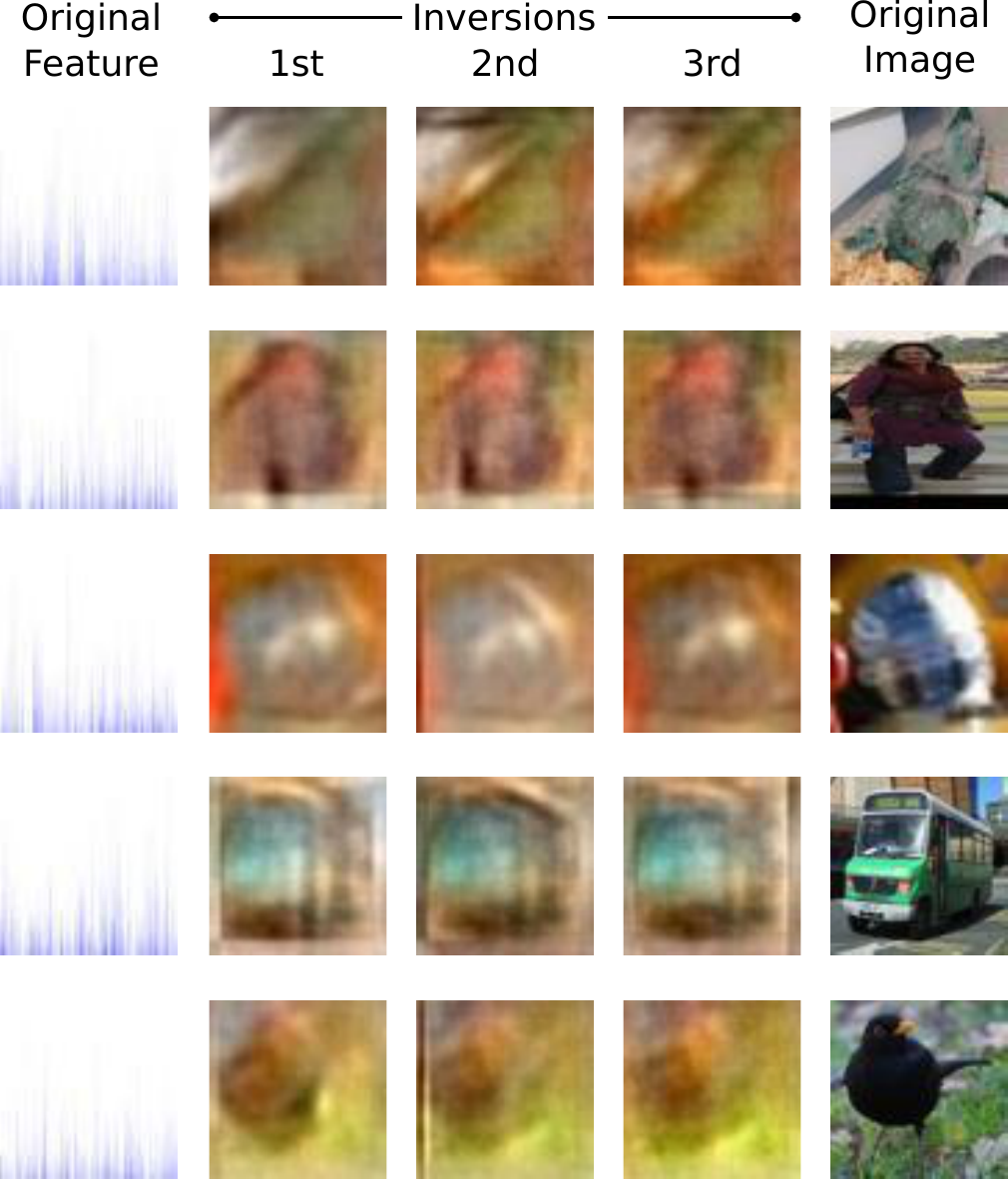}
}
\vspace{1em}
\subfloat[Nudged Dict]{
\includegraphics[width=0.47\linewidth]{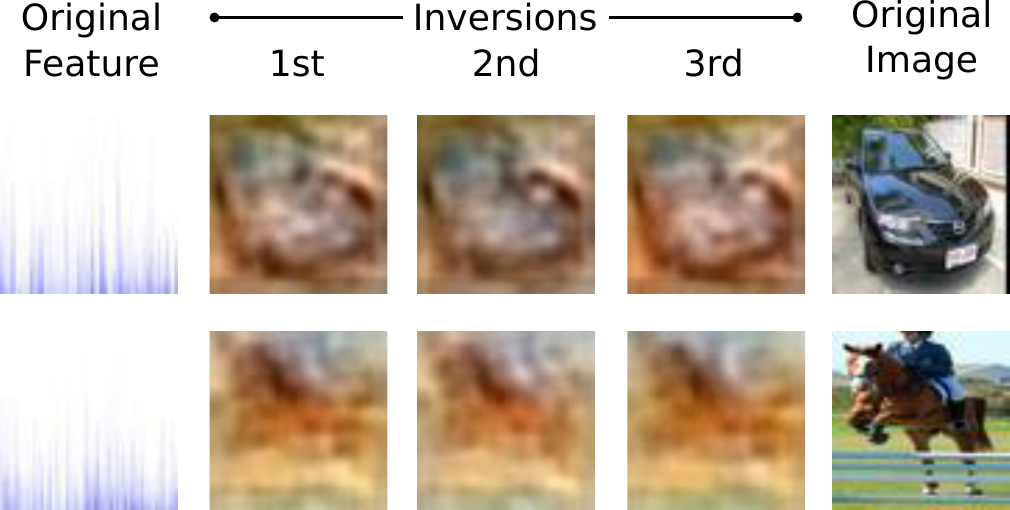}
}
\hspace{0.01\linewidth}
\subfloat[Subset Dict]{
\includegraphics[width=0.47\linewidth]{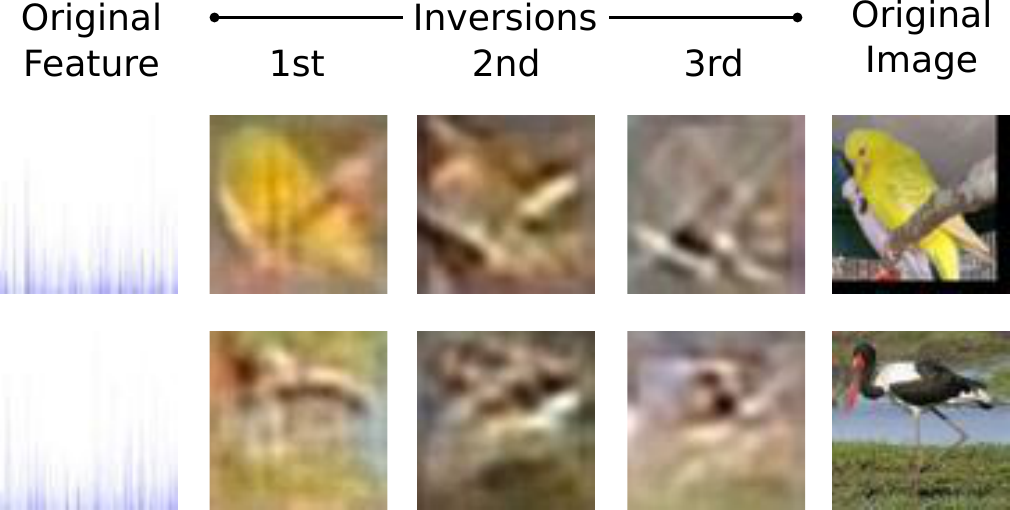}
}

\caption{We show the first three inversions for a few patches from our testing
set. Notice how the color (a) and edge (b) variants of our method tend to
produce different inversions. The baselines tend to either similar in image
space (c) or do not match well in feature space (d). Best viewed on screen. }

\label{fig:qual}
\end{figure}

\begin{figure}[t]
\centering
\includegraphics[width=0.48\linewidth]{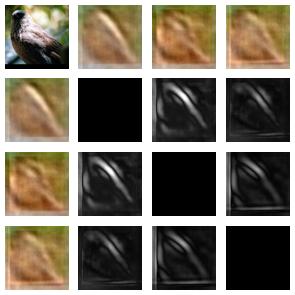}
\hspace{0.02\linewidth}
\includegraphics[width=0.48\linewidth]{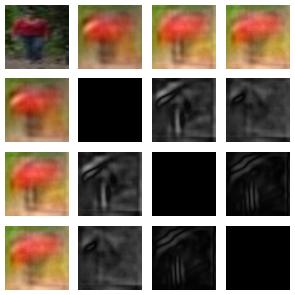}
%\\
%\vspace{0.02\linewidth}
%\includegraphics[width=0.48\linewidth]{figs/qual1/edge-matrix-3.jpg}
%\hspace{0.02\linewidth}
%\includegraphics[width=0.48\linewidth]{figs/qual1/edge-matrix-4.jpg}
\caption{The edge affinity can often result in subtle differences. Above, \changed{we show a difference matrix} between the first three inversions that highlights differences between
all pairs of a few inversions from one CNN feature. The margins show the inversions, and the inner black squares show the absolute difference. White means larger difference. Notice that our algorithm is able to recover inversions with shifts of gradients.}
\label{fig:edge}
\end{figure}

\subsection{Qualitative Results}
We first look at a few qualitative results for our multiple feature
inversions. Figure \ref{fig:qual} shows a few examples for both our method (top
rows) and the baselines (bottom rows). The 1st column shows the result of
a paired dictionary on CNN features, while the 2nd and 3rd show the
additional inversions that our method finds. While the results are blurred,
they do tend to resemble the original image in rough shape and color. 

The color affinity in Figure \ref{fig:qual}a is often able to produce inversions
that vary slightly in color. Notice how the cat and the floor are changing
slightly in hue, and the grass the bird is standing on is varying slightly. The
edge affinity in Figure \ref{fig:qual}b can occasionally generate inversions with
different edges, although the differences can be subtle. To better show the differences
with the edge affinity, we visualize a difference matrix in Figure \ref{fig:edge}.
Notice how the edges of the bird and person shift between each inversion.

\begin{figure}[t]
\centering
\includegraphics[width=0.8\linewidth]{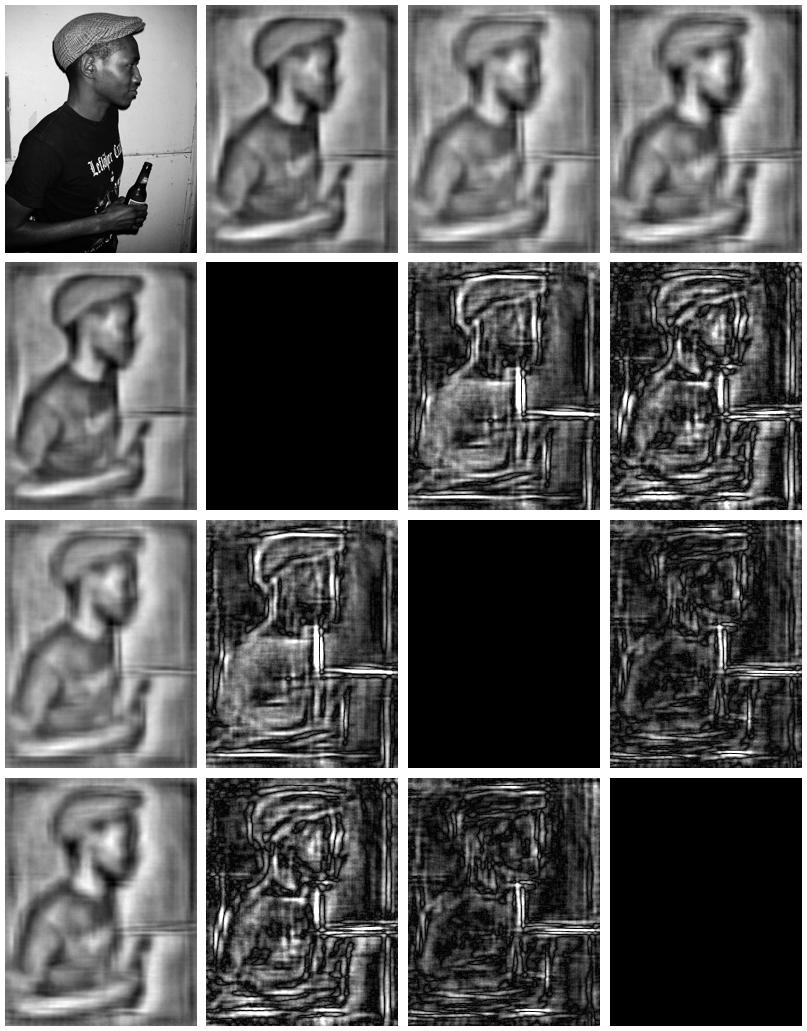}
\caption{\changed{The block-wise histograms of HOG allow for gradients in the
image to shift up to their bin size without affecting the feature descriptor.
By using our visualization algorithm with the edge affinity matrix, we can
recover multiple HOG inversions that differ by edges subtly shifting. Above, we
show a difference matrix between the first three inversions for a downsampled
image of a man shown in the top left corner. Notice the vertical gradient in
the background shifts between the inversions, and the man's head move
slightly.}}
\label{fig:multiple-hog}
\end{figure}

The baselines tend to either produce nearly identical inversions or inversions
that do not match well in feature space. Nudged dictionaries in Figure
\ref{fig:qual}c frequently retrieves inversions that look nearly identical.
Subset dictionaries in Figure \ref{fig:qual}d recovers different inversions,
but the inversions do not match in feature space, likely because this baseline
operates over a subset of the basis elements.

\changed{
Although HOG is not as invariant to visual transformations as deep features, we can
still recover multiple inversions from a HOG descriptor. The block-wise histograms of HOG allow for
gradients in the image to shift up to their bin size without affecting the feature descriptor. Figure \ref{fig:multiple-hog}
shows multiple inversions from a HOG descriptor of a man where the person shifts slightly between each inversion.
}

\subsection{Quantitative Results}

We wish to quantify how well our inversions
trade off matching in feature space versus having diversity in image space. To
evaluate this, we calculated Euclidean distance between the features of the
first and second inversions from each method, $||\phi(x_1) -
\phi(x_2)||_2$, and compared it to the Euclidean distance of the inversions in
Lab image space, $||L(x_1) - L(x_2)||_2$ where $L(\cdot)$ is the Lab
colorspace transformation.\footnote{We chose Lab because Euclidean distance in this
space is known to be perceptually uniform \citep{jain1989fundamentals}, which
we suspect better matches human interpretation.} We consider one inversion
algorithm to be better than another method if, for the same distance in feature space, the
image distance is larger.

\begin{figure}[t]
\centering
\includegraphics[trim=15em 1em 15em 1em,clip,width=\linewidth]{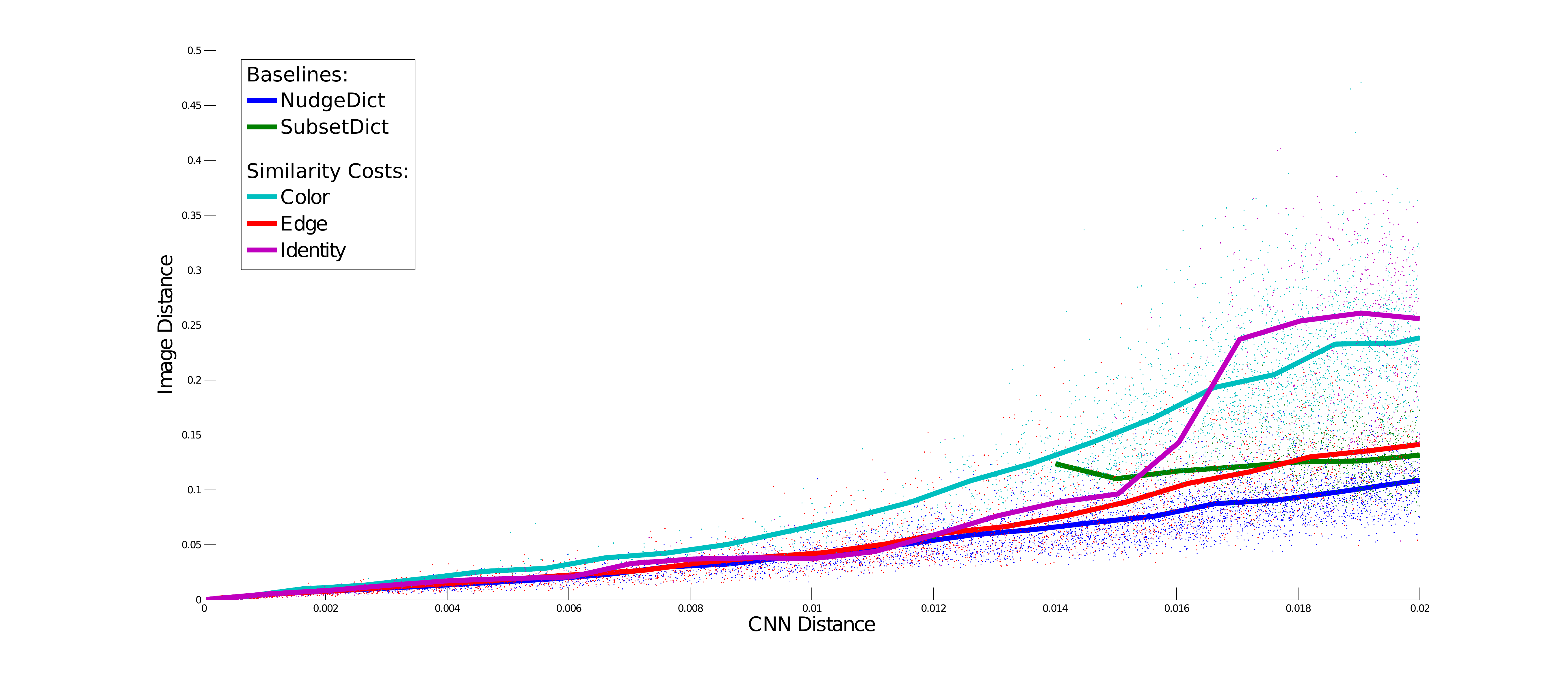}

\caption{We evaluate the
performance of our multiple inversion algorithm. The horizontal axis is the
Euclidean distance between the first and second inversion in CNN space and the
vertical axis is the distance of the same inversions in Lab colorspace. 
\changed{This plot suggests that incorporating diversity costs into the inversion are 
able to produce more diverse multiple visualizations for the same reconstruction error.}
Thick lines show the median image distance for a given feature distance.}

\label{fig:eval1}
\end{figure}

We show a scatter plot of this metric in
Figure \ref{fig:eval1} for our method with different similarity costs. The thick
lines show the median image distance for a given feature distance. The overall
trend suggests that our method produces more diverse images for the same
distance in feature space. Setting the affinity matrix $A$ to perform color
averaging produces the most image variation for CNN features in order to keep
the feature space accuracy small. The baselines in general do not perform as
well, and baseline with subset dictionaries struggles to even match in feature space, causing the
green line to abruptly start in the middle of the plot. The edge affinity
produces inversions that tend to be more diverse than baselines, although this effect
is best seen qualitatively in the next section.

We consider a second evaluation metric designed
to determine how well our inversions match the original features. Since
distances in a feature space are unscaled, they can be difficult to interpret, so we use a
normalized metric. We calculate the ratio of distances that the inversions make to the original
feature: $r = \frac{||\phi(x_2) - f||_2}{||\phi(x_1) - f||_2}$ where $f$ is the original
feature and $x_1$ and $x_2$ are the first and second inversions. A value of $r = 1$ implies
the second inversion is just as close to $f$ as the first. We then compare the ratio $r$ to 
the Lab distance in image space.

\begin{figure}[t]
\centering
\subfloat[Color]{\includegraphics[width=0.32\linewidth]{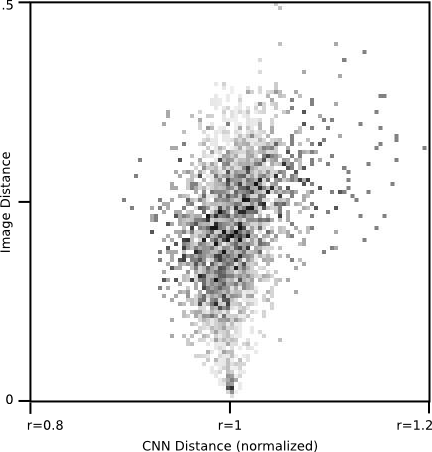}}
\hspace{.01em}
\subfloat[Identity]{\includegraphics[width=0.32\linewidth]{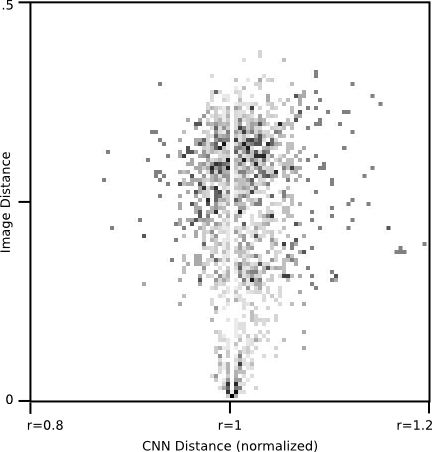}}
\hspace{.01em}
\subfloat[Edge]{\includegraphics[width=0.32\linewidth]{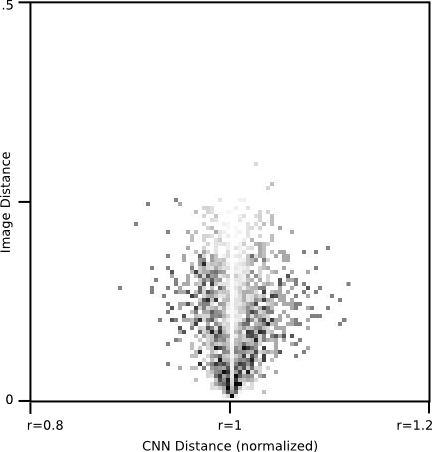}}
\\
\subfloat[Nudged Dict]{\includegraphics[width=0.32\linewidth]{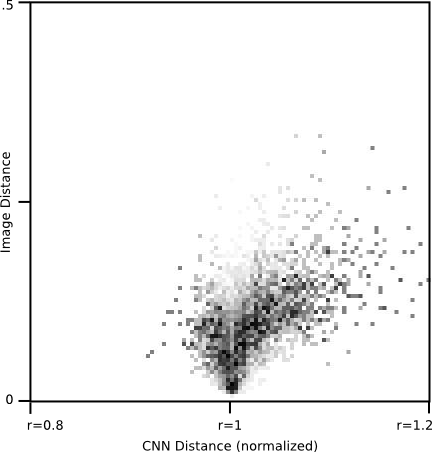}}
\hspace{.01em}
\subfloat[Subset Dict]{\includegraphics[width=0.32\linewidth]{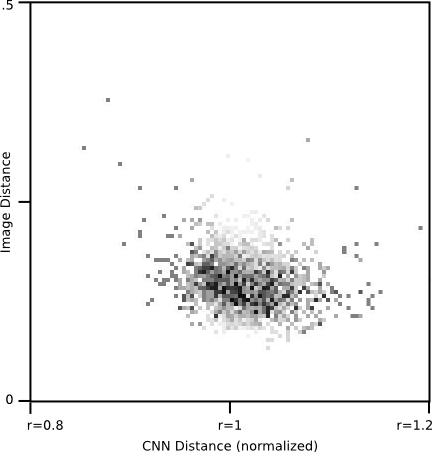}}

\caption{We show density maps that visualize image
distance versus the ratio distances in feature space: $r = \frac{||\phi(x_2)-f||_2}{||\phi(x_1)-f||_2}$. 
A value of $r = 1$ means that the two inversions are the same distance from the original feature. Black means most dense
and white is zero density. Our results suggest that our method with the affinity matrix set to color averaging produces more diverse
visualizations for the same $r$ value.}

\label{fig:eval2}
\end{figure}

We show results for our second metric in
Figure \ref{fig:eval2} as a density map comparing image distance and the ratio of
distances in feature space. Black is a higher density and implies that the
method produces inversions in that region more frequently. This experiment
shows that for the same ratio $r$, our approach tends to produce more diverse
inversions when affinity is set to color averaging. Baselines frequently performed poorly,
and struggled to generate diverse images that are close in feature space.

%One limitation of our visualization is that it tends to be blurry. We suspect
%that increasing the dictionary size $K$ will yield more crisp visualizations at
%the expense of speed. If we increase the dictionary size without bound, then
%our approach will approach a nearest neighbor based visualization, but such an
%approach would be slow.  We believe that visualizations that are interactive
%are more useful for computer vision researchers today. We expect that advances
%in fast sparse coding methods will be impactful in feature visualization. 

\section{Understanding Object Detectors}

\changed{While the goal of this paper is to visualize object detection
features, in this section we will use our visualizations to inspect the
behavior of object detection systems. Due to our budget for experiments, we focus
on HOG features.}

\subsection{HOGgles}

\begin{figure}
\centering
\subfloat[Human Vision]{
\includegraphics[height=18em]{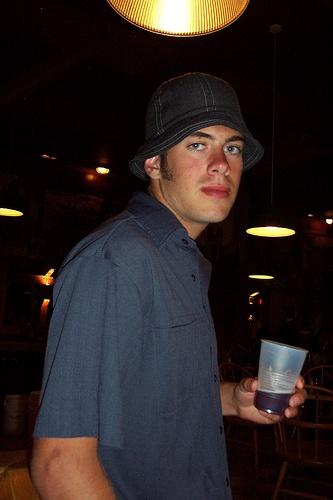}
\label{fig:seeintodarkA}
}
\subfloat[HOG Vision]{
\includegraphics[height=18em]{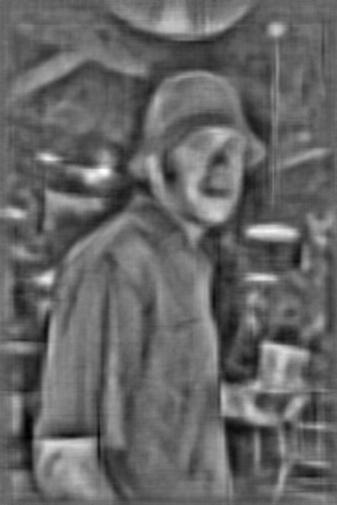}
\label{fig:seeintodarkB}
}

\caption{Feature inversion reveals the world that object detectors see. The left
shows a man standing in a dark room. If we compute HOG on this
image and invert it, the previously dark scene behind the man emerges. Notice
the wall structure, the lamp post, and the chair in the bottom right hand
corner.}

\label{fig:seeintodark}

\end{figure}

Our visualizations reveal that the world that features see is slightly
different from the world that the human eye perceives. Figure
\ref{fig:seeintodarkA} shows a normal photograph of a man standing in a dark
room, but Figure \ref{fig:seeintodarkB} shows how HOG features see the same
man. Since HOG is invariant to illumination changes and amplifies gradients,
the background of the scene, normally invisible to the human eye, materializes
in our visualization. 

In order to understand how this clutter affects object detection, we visualized
the features of some of the top false alarms from the Felzenszwalb et al.\
object detection system \citep{felzenszwalb2010object} when applied to the
PASCAL VOC 2007 test set.  Figure \ref{fig:topdets} shows our visualizations of
the features of the top false alarms. Notice how the false alarms look very
similar to true positives. While there are many different types of detector
errors, this result suggests that these particular failures are due to
limitations of HOG, and consequently, even if we develop better learning
algorithms or use larger datasets, these will false alarms will likely persist.

\begin{figure*}
\centering
\includegraphics[width=\linewidth]{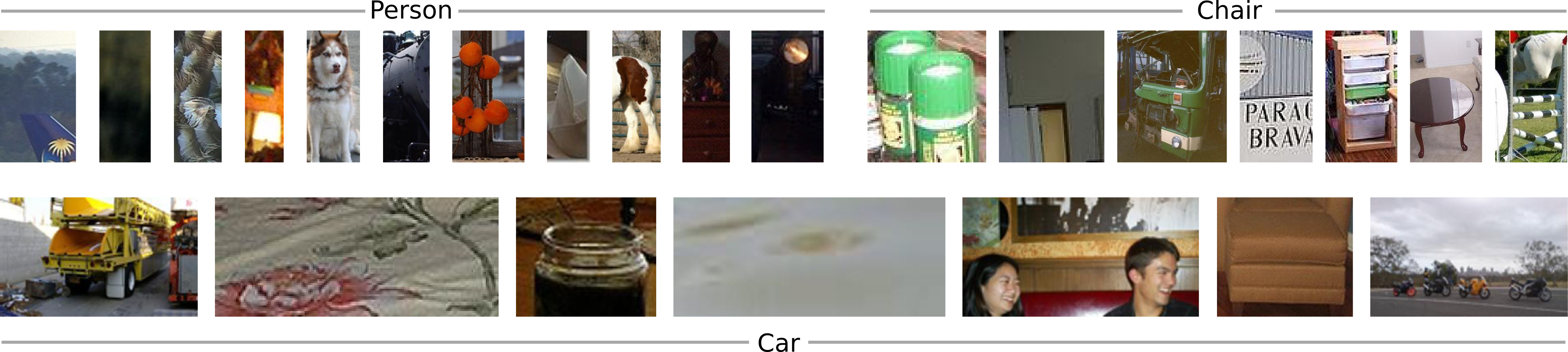}
\caption{We show the original RGB patches that correspond to the
visualizations from Figure \ref{fig:topdets}.  We print the original patches on
a separate page to highlight how the inverses of false positives look
like true positives. We recommend comparing this figure side-by-side with
Figure \ref{fig:topdets}.} \label{fig:topdetsrgb}
\end{figure*}

Figure \ref{fig:topdetsrgb} shows the corresponding RGB image patches for the
false positives discussed above. Notice how when we view these detections in
image space, all of the false alarms are difficult to explain. Why do chair
detectors fire on buses, or people detectors on cherries? By
visualizing the detections in feature space, we discovered that the learning
algorithm made reasonable failures since the features are deceptively
similar to true positives.

\subsection{Human+HOG Detectors}

Although HOG features are designed for machines, how well do humans see in HOG
space? If we could quantify human vision on the HOG feature space, we could get
insights into the performance of HOG with a perfect learning algorithm
(people). Inspired by Parikh and Zitnick's methodology
\citep{parikh2011human,parikh2010role}, we conducted a large human study where
we had Amazon Mechanical Turk participants act as sliding window HOG based object
detectors.

We built an online interface for humans to look at HOG visualizations of window
patches at the same resolution as DPM. We instructed participants to either classify
a HOG visualization as a positive example or a negative example for a category.
By averaging over multiple people (we used 25 people per window), we obtain a
real value score for a HOG patch. To build our dataset, we sampled top
detections from DPM on the
PASCAL VOC 2007 dataset for a few categories. Our dataset consisted of around $5,000$
windows per category and around $20\%$ were true positives.

\begin{figure}
\centering
\includegraphics[width=0.48\linewidth]{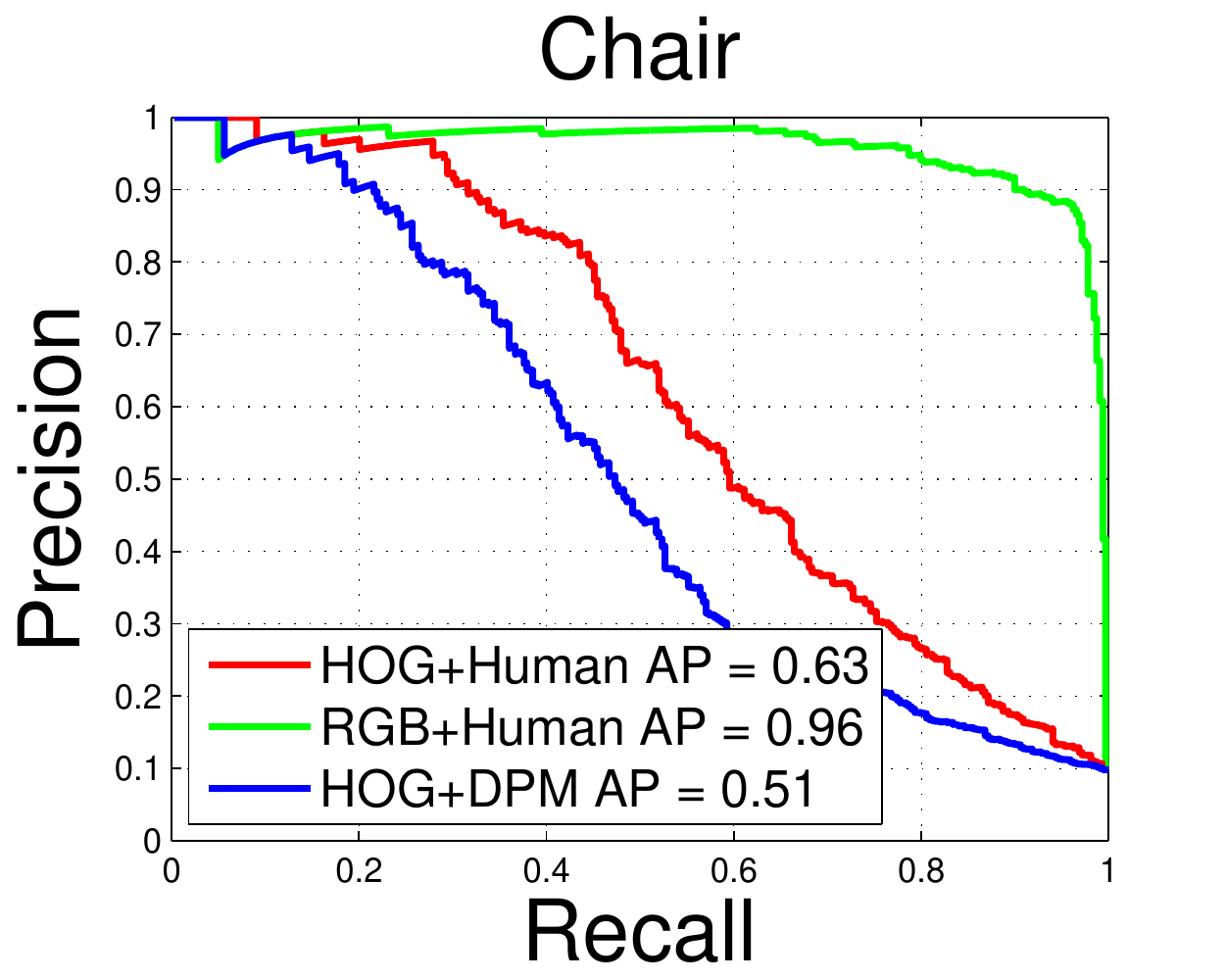}
\includegraphics[width=0.48\linewidth]{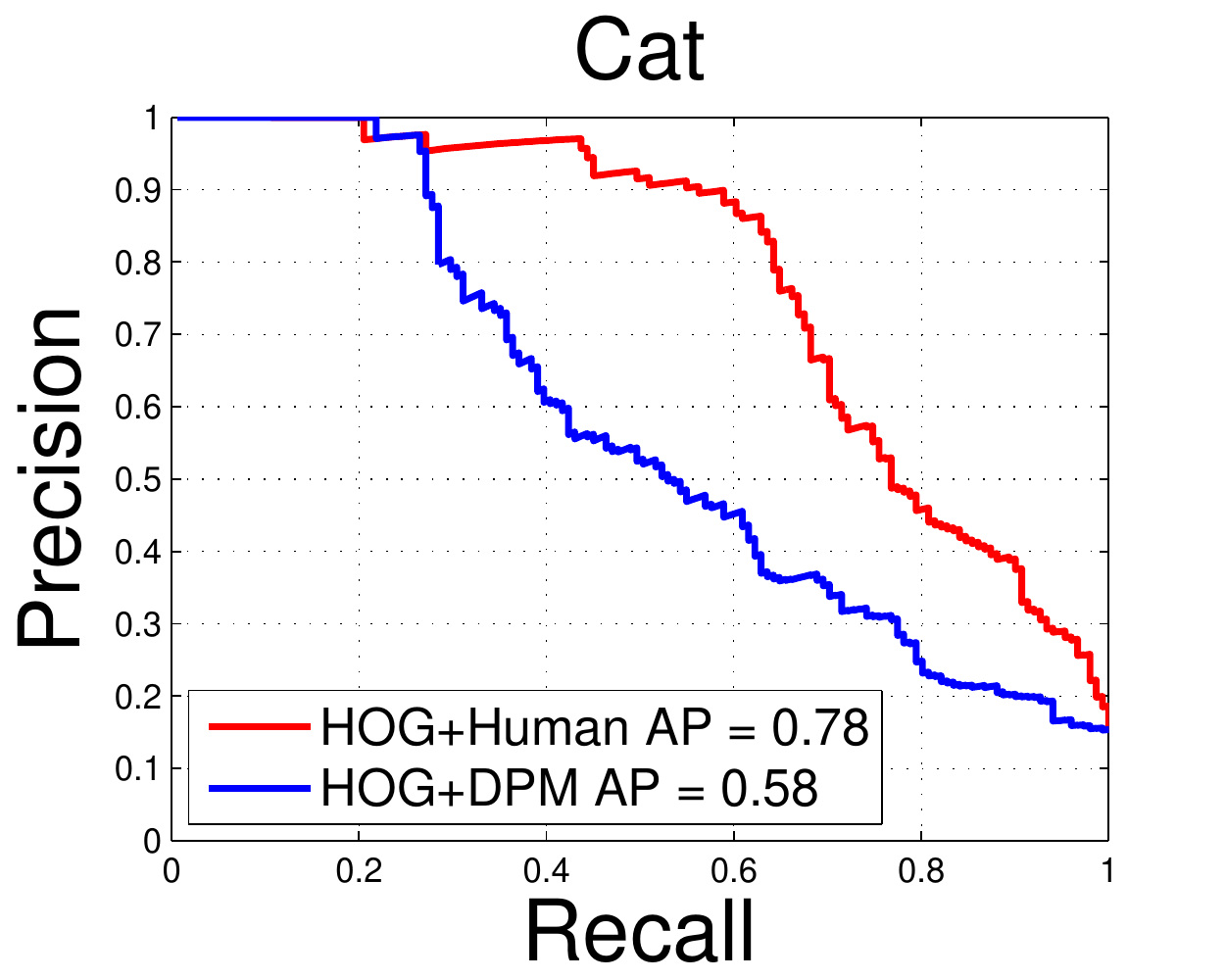}
\includegraphics[width=0.48\linewidth]{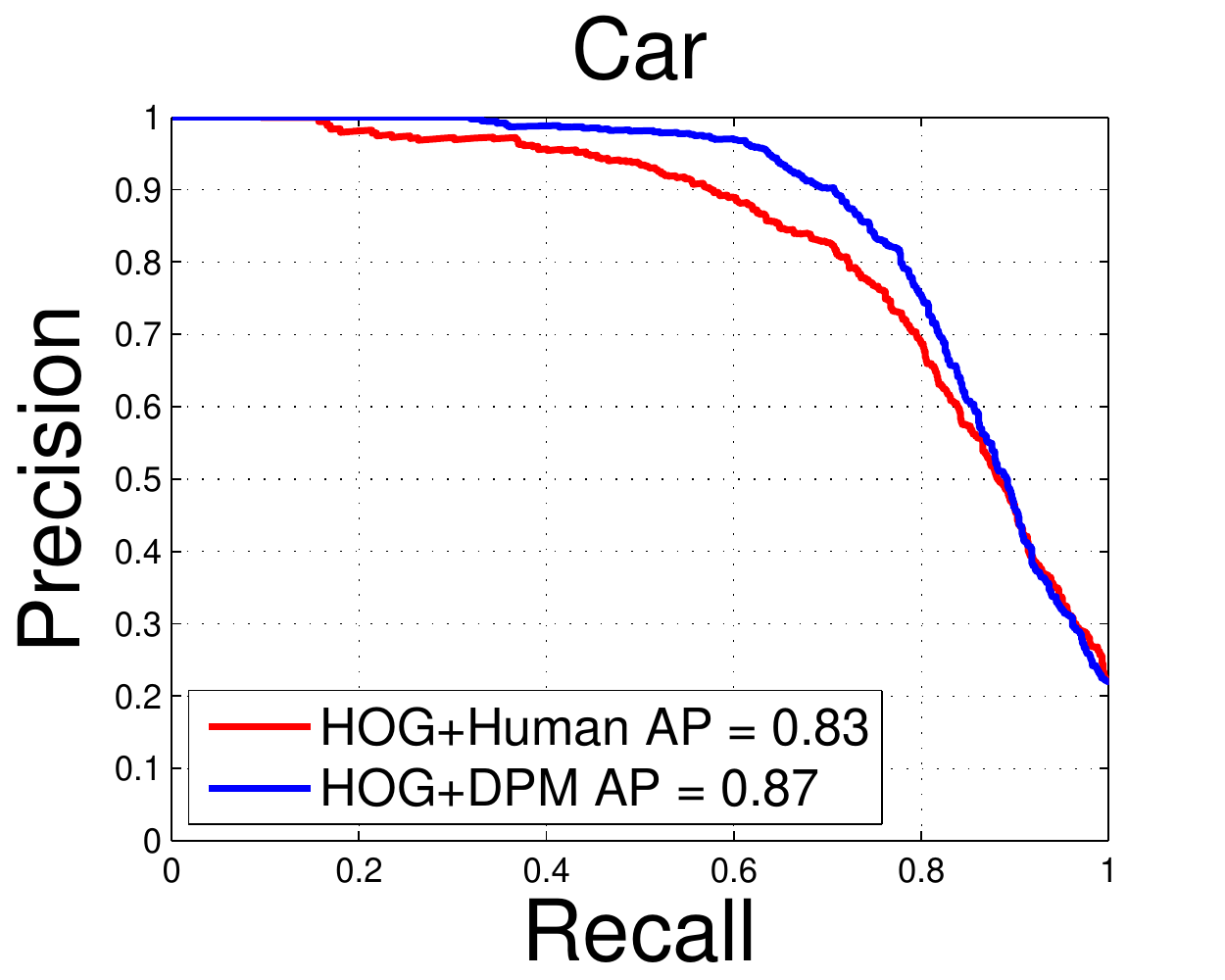}
\includegraphics[width=0.48\linewidth]{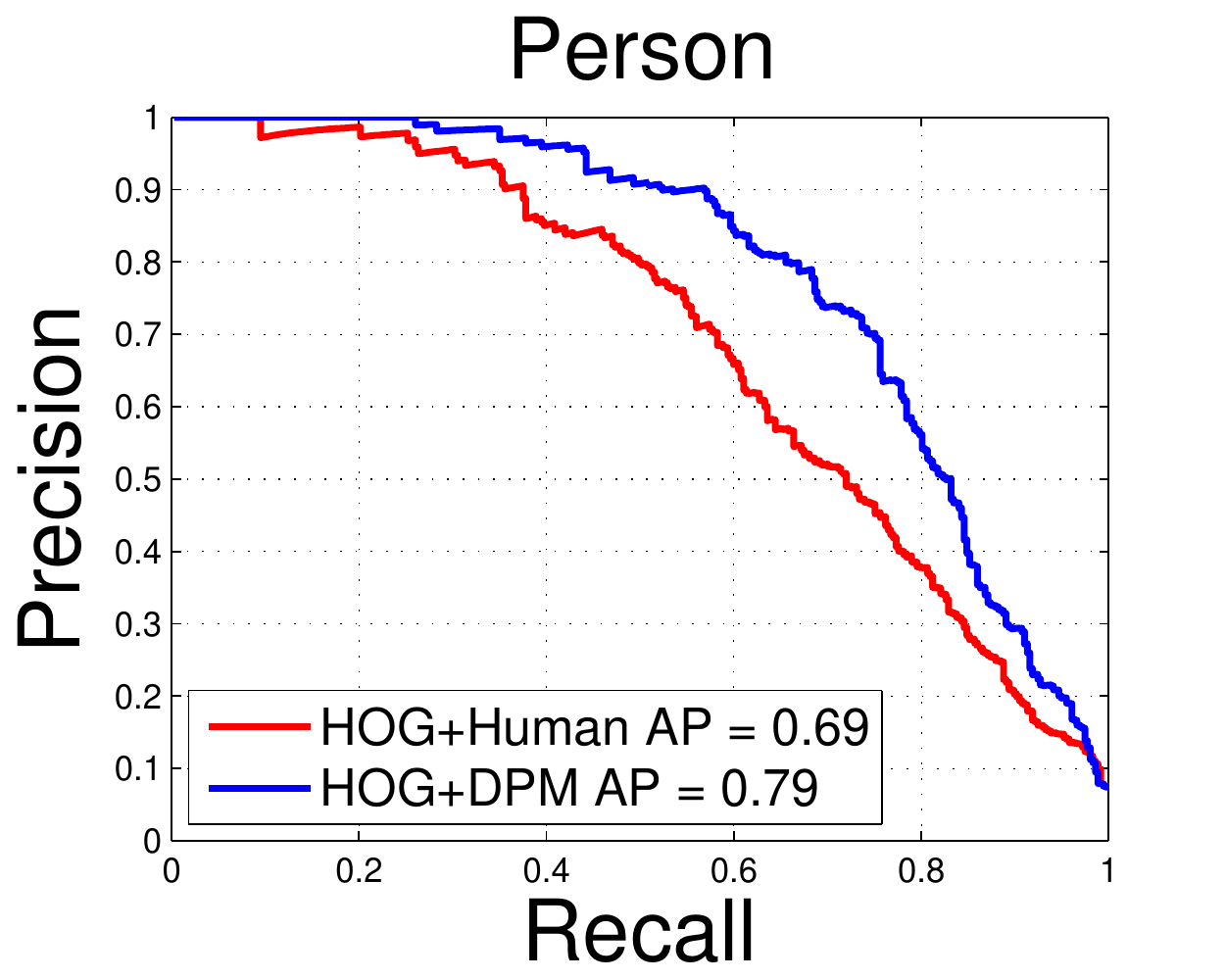}

\caption{By instructing multiple human subjects to classify the visualizations, we show
performance results with an ideal learning algorithm (i.e., humans) on the HOG
feature space. Please see text for details.}

\label{fig:hoggles}

\end{figure}

Figure \ref{fig:hoggles} shows precision recall curves for the Human + HOG based
object detector. In most cases, human subjects classifying HOG visualizations
were able to rank sliding windows with either the same accuracy or better than
DPM. Humans tied DPM for recognizing cars, suggesting that performance may be
saturated for car detection on HOG.  Humans were slightly superior to DPM for
chairs, although performance might be nearing saturation soon. There appears to
be the most potential for improvement for detecting cats with HOG. Subjects
performed slightly worst than DPM for detecting people, but we believe this is
the case because humans tend to be good at fabricating people in abstract
drawings.

We then repeated the same experiment as above on chairs except we instructed
users to classify the original RGB patch instead of the HOG visualization. As
expected, humans have near perfect accuracy at detecting chairs with
RGB sliding windows. The performance gap between the Human+HOG detector and Human+RGB detector
demonstrates the amount of information that HOG features discard. 

%\begin{figure}
%\includegraphics[width=0.5\linewidth]{figs/chair-hoggles-corr.pdf}\includegraphics[width=0.5\linewidth]{figs/cat-hoggles-corr.pdf}
%
%\includegraphics[width=0.5\linewidth]{figs/car-hoggles-corr.pdf}\includegraphics[width=0.5\linewidth]{figs/person-hoggles-corr.pdf}
%
%\caption{We plot the correlation coefficient to DPM's ranking of detections by
%the human detector versus the percent of detections returned. Since humans
%cannot accurately score how the liklihood an object is not a category, the
%detections become decorrelated if we retrieve the entire dataset. However,
%there is statistically significant correlation if we only consider the top 20\%
%high scoring windows. Therefore, humans looking at HOG visualizations produce
%scores similar to DPM.}
%
%\label{fig:hoggles-correlation}
%
%\end{figure}

\begin{figure*}
\centering
\includegraphics[height=6.7em]{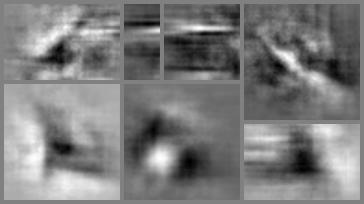}\hspace{0.5em}
\includegraphics[height=6.7em]{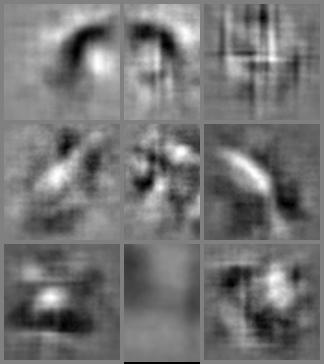}\hspace{0.5em}
\includegraphics[height=6.7em]{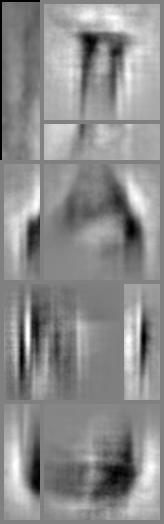}\hspace{0.5em}
\includegraphics[height=6.7em]{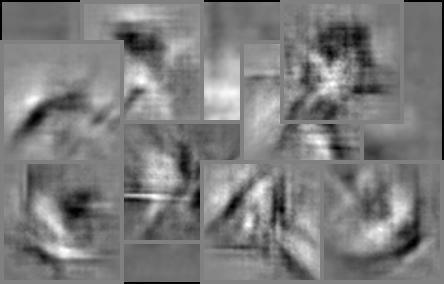}\hspace{0.5em}
\includegraphics[height=6.7em]{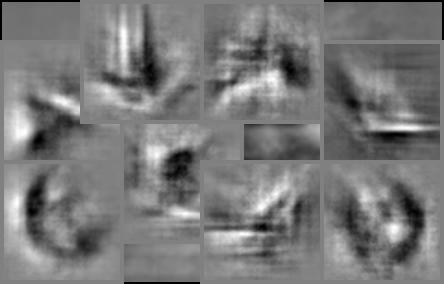}\hspace{0.5em}
\includegraphics[height=6.7em]{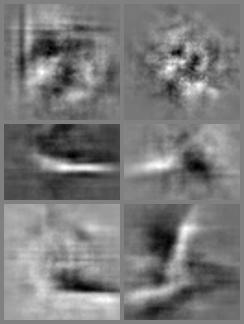}\hspace{0.1em}\includegraphics[height=6.7em]{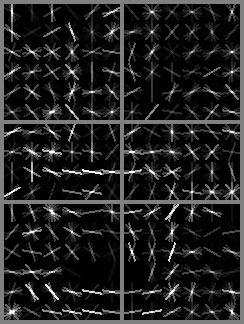} \\
\vspace{0.5em}
\includegraphics[height=6.7em]{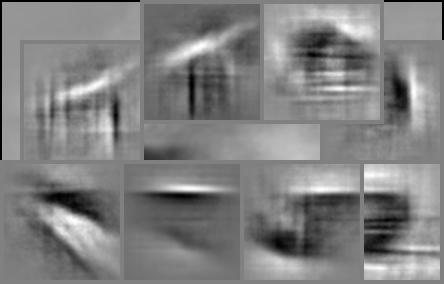}\hspace{0.5em}
\includegraphics[height=6.7em]{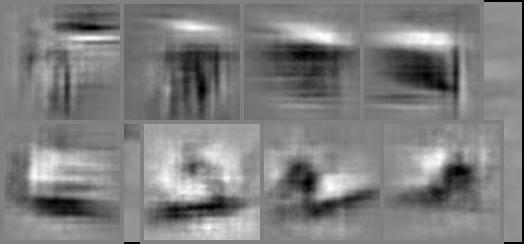}\hspace{0.5em}
\includegraphics[height=6.7em]{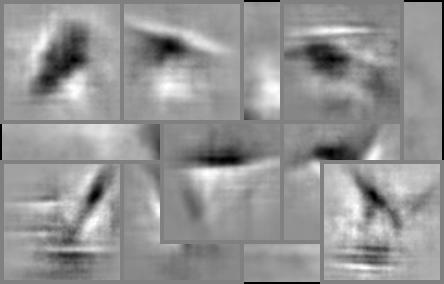}\hspace{0.5em}
\includegraphics[height=6.7em]{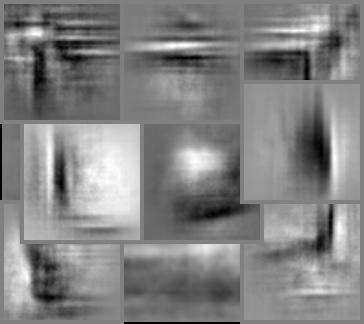}\hspace{0.5em}
\includegraphics[height=6.7em]{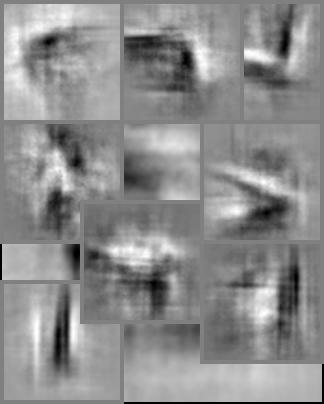}\hspace{0.1em}\includegraphics[height=6.7em]{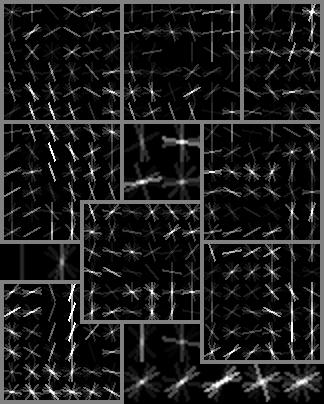}

\caption{We visualize a few deformable parts models trained with
\citep{felzenszwalb2010object}. Notice the structure that emerges with our
visualization.  First row: car, person, bottle, bicycle, motorbike, potted
plant. Second row: train, bus, horse, television, chair. For the right most
visualizations, we also included the HOG glyph. Our visualizations
tend to reveal more detail than the glyph.}

\label{fig:prototypes}

\end{figure*}

Our experiments suggest that there is still some performance left to be
squeezed out of HOG. However, DPM is likely operating very close to the
performance limit of HOG. Since humans are the ideal learning agent and they
still had trouble detecting objects in HOG space, HOG may be too lossy of a
descriptor for high performance object detection.  If we wish to significantly
advance the state-of-the-art in recognition, we suspect focusing effort on
building better features that capture finer details as well as higher level
information will lead to substantial performance improvements in object
detection. Indeed, recent advances in object recognition have been driven
by learning with richer features \citep{girshick2013rich}.

\subsection{Model Visualization}

We found our algorithms are also useful for visualizing the learned models of an
object detector.  Figure \ref{fig:prototypes} visualizes the root templates and
the parts from \citep{felzenszwalb2010object} by inverting the
positive components of the learned weights. These visualizations provide hints on which
gradients the learning found discriminative.  Notice the detailed structure
that emerges from our visualization that is not apparent in the HOG glyph. Often, one can recognize the category of the detector by only looking at the
visualizations.

\section{Conclusion}

We believe visualizations can be a powerful tool for understanding object
detection systems and advancing research in computer vision. To this end, this
paper presented and evaluated several algorithms to visualize object detection
features. We hope more intuitive visualizations will prove useful for the
community.

\emph{Acknowledgments:} We thank the CSAIL Vision Group
for many important discussions. Funding was provided by a NSF GRFP to CV, a
Facebook fellowship to AK, and a Google research award, ONR MURI N000141010933
and NSF Career Award No. 0747120 to AT.

\bibliographystyle{spbasic}
\bibliography{main}

\end{document}